\documentclass[journal, twocolumn]{IEEEtran}
\usepackage{setspace,cite}
\usepackage{graphicx}
\usepackage{epstopdf}
\usepackage{caption}
\usepackage{epsfig} 
\usepackage{amsmath} 
\usepackage{amssymb} 
\usepackage{amsthm}
\usepackage[dvipsnames]{xcolor}
\usepackage{float}
\usepackage{subcaption}
\usepackage{verbatim}  
\usepackage{comment}  
\usepackage{multirow}
\newtheorem{theorem}{Theorem}
\newtheorem{proposition}{Proposition}

\newtheorem{remark}{Remark}
\newtheorem{definition}{Definition}[section]

\usepackage[ colorlinks = true,
linkcolor = blue,
urlcolor = blue,
citecolor = red,
anchorcolor = green,
]{hyperref}

\allowdisplaybreaks[2]

\usepackage{hyperref}
\usepackage[english]{babel}
\usepackage[T1]{fontenc}
\usepackage{amsfonts}
\usepackage{algorithm}	
\usepackage{enumitem}
\usepackage{tabularx}
\usepackage{multirow}
\usepackage{colortbl} 
\usepackage{arydshln} 

\usepackage{algorithm}	
\usepackage[noend]{algpseudocode}
\usepackage{booktabs}
\usepackage{amssymb}
\usepackage{amsmath}
\usepackage{graphics} 
\usepackage{graphicx, color}
\usepackage{tikz}
\usetikzlibrary{shapes,arrows}
\usepackage{verbatim}
\usepackage{subcaption}
\usepackage[flushleft]{threeparttable}
\newcommand{\diag}{\mathop{\rm diag}}

\usetikzlibrary{arrows,positioning, decorations.pathreplacing,decorations.pathmorphing}
\tikzset{
    >=stealth',
    punkt/.style={
           rectangle,
           rounded corners,
           draw=black, very thick,
           text width=6.5em,
           minimum height=2em,
           text centered},
    pil/.style={
           ->,
           thick,
           shorten <=2pt,
           shorten >=2pt,},
    decoration={brace},
  	tuborg/.style={decorate},
	tubnode/.style={midway, right=2pt}
}
\pgfdeclarelayer{background}
\pgfdeclarelayer{foreground}
\pgfsetlayers{background,main,foreground}
\tikzstyle{block} = [draw, text width=4em, text centered, fill=blue!20, rectangle,
    minimum height=3em, minimum width=4em]
\tikzstyle{sum} = [draw, fill=blue!20, circle, node distance=0.01cm]
\tikzstyle{input} = [coordinate]
\tikzstyle{output} = [coordinate]
\tikzstyle{pinstyle} = [pin edge={to-,thin,black}]
\tikzstyle{blockAtt} = [draw, fill=red!20, rectangle,
    minimum height=3em, minimum width=4em,text width=4em, text centered]
\tikzstyle{annot} = [text width= 3em, text centered]

\DeclareMathOperator*{\argmax}{\arg\!\max}
\DeclareMathOperator*{\argmin}{\arg\!\min}

\def \extended {1}

\author{Peter Kairouz, Jiachun Liao, Chong Huang, Maunil Vyas, Monica Welfert, Lalitha Sankar
\thanks{This work is supported in part by the National Science Foundation under grants CCF-1350914, CIF-1815361, CIF-1901243, SaTC-2031799, DMS-2134256, and CIF-2007688. The associate editor coordinating the review of this manuscript and approving it for publication was Prof. Matthieu Bloch. (Peter Kairouz and Jiachun Liao are co-first authors.) (Corresponding author: Monica Welfert.)}
\thanks{M. Welfert and L. Sankar are with the School of Electrical, Computer, and Energy Engineering, Arizona State University, Tempe, AZ 85281 USA, as were J. Liao, C. Huang, and M. Vyas at the time the work was done (email: \{mwelfert, lsankar, jliao15, chuang83, mrvyas\}@asu.edu). P. Kairouz is with Google Research, Mountain View, CA 94043.  (email: kairouz@google.com).}
\thanks{This paper has supplementary downloadable material available at \url{https://doi.org/}, provided by the author.}
\thanks{Digital  Object Identifier  }}

\begin{document}
\title{Generating Fair Universal Representations using Adversarial Models}

\maketitle
\graphicspath{{./}}

\begin{abstract}
We present a data-driven framework for learning \textit{{fair universal representations}} (FUR) that guarantee statistical fairness for {any} learning task that may not be known \textit{a priori}. 
Our framework leverages recent advances in adversarial learning to allow a data holder to learn representations in which a set of sensitive attributes are decoupled from the rest of the dataset. We formulate this as a constrained minimax game between an encoder and an adversary where the constraint ensures a measure of usefulness (utility) of the representation. The resulting problem is that of censoring, i.e., finding a representation that is least informative about the sensitive attributes given a utility constraint. For appropriately chosen adversarial loss functions, our censoring framework precisely clarifies the optimal adversarial strategy against strong information-theoretic adversaries; it also achieves the fairness measure of demographic parity for the resulting constrained representations. We evaluate the performance of our proposed framework on both synthetic and publicly available datasets.
For these datasets, we use two tradeoff measures: censoring vs. representation fidelity and fairness vs. utility for downstream tasks, to amply demonstrate that multiple sensitive features can be effectively censored even as the resulting fair representations ensure accuracy for multiple downstream tasks.

\end{abstract}

\vspace{-0.15cm}

\section{Introduction}

The use of data-driven machine learning (ML) has recently seen unprecedented success in a variety of automated decision-making systems including facial recognition, natural language processing, mortgage lending, and even parole prediction. The success of these approaches hinges on the availability of large datasets that often include sensitive personal information. 
It has been shown that models learned from such datasets can inherit 
societal bias and discrimination patterns \cite{ladd1998evidence,pedreshi2008discrimination} and learn sensitive features even when they are not explicitly used during training \cite{song2019overlearning}. 
Concerns about the fairness, bias, and privacy of learning algorithms have led to a growing body of research focused on both defining meaningful fairness measures and designing algorithms with such guarantees. 


A key challenge in algorithmic fairness is the quantification of disparate treatment and impact -- legal notions developed to ensure that societal decisions neither hinder nor discriminate against specific groups. Addressing this has broadly lead to two classes of measures: (i) group fairness measures which require similar outcomes for all groups \cite{certifying_feldman2015}; (ii) individual fairness measures which require treating similar individuals 
similarly \cite{dwork2012fairness}. Approaches combining both fairness requirements have also been considered \cite{Calmon2018data,kearns2018preventing}. In the context of supervised learning of intended tasks (our setting here), two key group fairness measures have emerged \cite{hardt2016equality}: (i) demographic parity (DemP) which requires predicted outcomes to be independent of the sensitive features, and (ii) equalized odds (EO) wherein such an independence holds only when conditioned on the true outcome. The EO measure was introduced to ensure accurate predictions within groups, a limitation of DemP \cite{chouldechova2017fair}.


{Three distinct approaches have been considered to enforce fairness in learning: 
in-processing, pre-processing, and post-processing. In-processing approaches are most commonly used in the supervised setting where the learning objective is known (e.g., \cite{dwork2012fairness,zhang2018mitigating}); the resulting trained model guarantees fairness for the specific objective. 
Pre-processing generally produces fair representations of data tuned for a chosen learning objective \cite{Madras_2018,edwards_Storkey2016,calmon2017optimized} while post-processing provides fairness by properly altering decision outputs \cite{hardt2016equality,hajian2015discrimination,wei20a-opt}. 
}

Recently, \textit{censoring} has emerged as a compelling pre-processing approach wherein protected features (e.g., race, gender, and their correlates) are actively decorrelated from the rest of the data to explicitly limit their effect on decisions. Censoring is inspired by information-theoretic privacy methods to limit leakage of sensitive features
\cite{hamm2017minimax,huang2017context,tripathy2017privacy,bertran2019adversarially,song2019overlearning} and can be achieved in practice using generative adversarial networks (GANs) \cite{goodfellow2014generative}. 
Thus far, censoring for fairness has largely focused on learning fair predictors \cite{edwards_Storkey2016,Madras_2018,zhang2018mitigating}.

\begin{figure}[ht]
\centering
\vspace{-0.1in}
\includegraphics[width=0.48\textwidth]{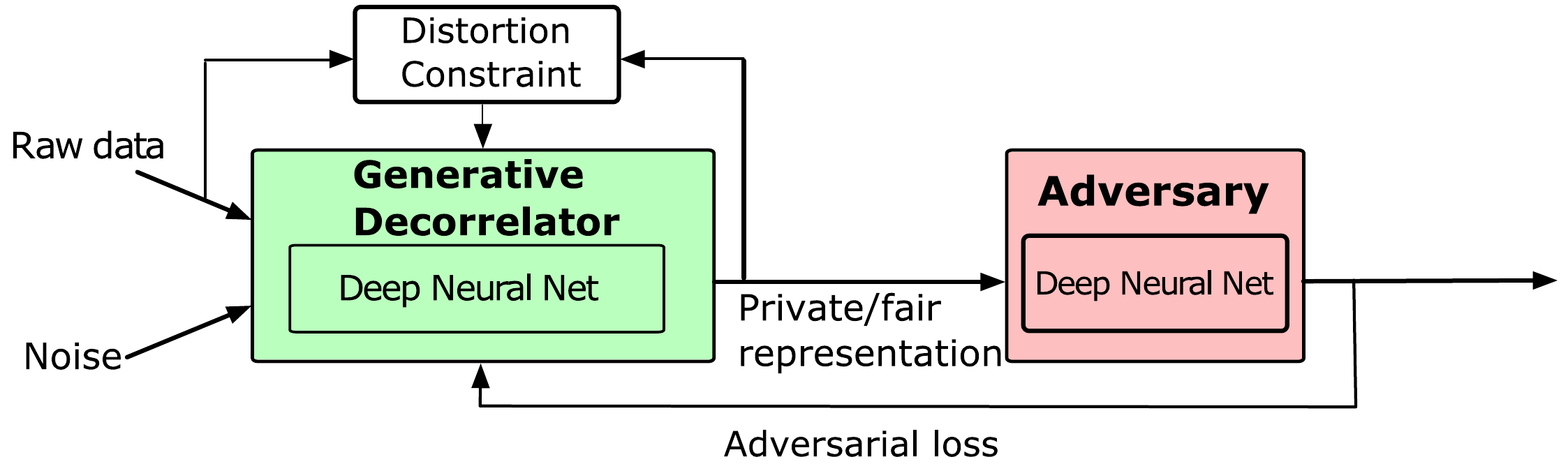}
\caption{Generative adversarial model for censoring/fairness.}
\label{fig:gap}
\end{figure}

\vspace{-0.05in}
\textbf{Our Contributions}: Taking a preprocessing approach, the main contribution of this work is to use censoring to generate fair representations that are \emph{universal}. These are representations from which the sensitive features have been actively decoupled and can be universally used for a variety of \emph{a priori} unknown learning tasks. We show that such fair universal representations (FURs) can assure DemP group fairness 
for all downstream predictions (from the data processing inequality). 
We now detail our contributions:
\begin{itemize}[leftmargin=*]
\item We present a framework for learning FURs as a \emph{constrained }minimax game between an encoder\footnote{referred to interchangeably throughout the paper as fair encoder or generator or decorrelator} and an adversary, where the encoder \emph{generates a noisy representation} of the original data, subject to a distortion constraint, 
to thwart an adversary that actively tries to infer the sensitive features (see Fig. \ref{fig:gap}).  {There has been recent work on using adversarial methods to generate \emph{transferable} fair representations~\cite{Madras_2018}; our universal FUR approach, while similar in philosophy, goes a step further by enforcing a \emph{hard} distortion constraint that allows better control of the learned representations, and therefore, better downstream utility guarantees. 
Algorithmically, we showcase how Lagrange penalty methods~\cite{lillo1993solving} can be leveraged to enforce the hard constraint in a GAN-setting\footnote{Recently TensorFlow updated its package to allow enforcing hard constraints \cite{TF-ConstOpt} using a similar approach.}.}
{\item Building on existing definitions of fair predictors, we formally define demographic parity for FRs.  We use censoring (of the sensitive features) to ensure DemP FRs and provide information-theoretic assurances on both. Censoring methods, used often for assuring information-theoretic privacy of sensitive features when releasing data (e.g., \cite{bertran2019adversarially,huang2017context}), can also ensure DemP fairness (relative to the sensitive feature). Building on this, we formally define censored representations (Definition \ref{Def:Censored_representation}) for the setting when adversaries are limited to practical ML models and loss functions. This has a broader value in auditing fairness/censoring guarantees.}
{\item Our prior work \cite{huang2017context} shows that the constrained FUR minimax game captures a range of adversarial actions through the choice of loss functions and identifies the corresponding information-theoretic optimal adversarial strategies. 
Focusing here on high-dimensional Gaussian mixture models with independent components, we present the optimal additive Gaussian noise distribution that minimizes the adversary's probability of detecting the mixture class (sensitive feature), i.e., the game-theoretic optimal under the strongest MAP (maximum \textit{a posteriori}) adversary. We then use normalized mutual information estimates to show that the empirical FUR framework, using neural networks and log-loss, performs very well relative to this game-theoretic optimal.}  
{\item Our most important contribution is in illustrating the utility of FURs for multiple publicly available datasets including the
	UCI Adult~\cite{kohavi1996scaling}, 
	UTKFace~\cite{zhifei2017cvpr},  
	GENKI~\cite{whitehill2012discriminately},
	and HAR~\cite{anguita2013public} datasets. 
	Our visual results demonstrate our success in creating high quality representations that increasingly erase the sensitive attributes with decreasing fidelity requirements. In contrast to state-of-the-art \cite{Madras_2018,edwards_Storkey2016,zhang2018mitigating}, our theoretical framework and experiments are the first to include non-binary sensitive attributes, multiple downstream tasks, as well as hard distortion constraints. Our results show that one can still learn high accuracy DemP (and even EO) fair classifiers from DemP FURs. In particular, via the UCI dataset, that is often used in fair ML analyses, we showcase the advantage of our approach relative to related approaches {(e.g., \cite{edwards_Storkey2016},  \cite{Madras_2018,zhang2018mitigating,Greager-Zemel_Flexibly-2019}).}
	Our results also straddle a wide range of values for the chosen fairness measure (DemP) and include perfect fairness, in contrast to the above works.}  
    
\end{itemize}

The remainder of our paper is organized as follows. We set up the problem and review known measures for fair predictors in Section \ref{sec:preliminary}. In Section  \ref{sec:model}, we formalize our FUR model, introduce definitions for censored and fair representations, and highlight the theoretical guarantees of this approach. In Section \ref{sec:gmm}, we present theoretical results for datasets modeled as multi-dimensional Gaussian mixtures. {Finally, we showcase the performance of the FUR framework on the UCI Adult, UTKFace, GENKI, and HAR datasets in Section \ref{sec:genkisim}. 
Proofs for the key results are in the appendix. All proofs that build on prior results in \cite{huang2017context} can be found in an extended version \cite{CFUR-LS}. 
Finally, details of the FUR architecture for all datasets are in the accompanying supplementary material.}
\vspace{-0.1in}
\section{Preliminaries}\label{sec:preliminary}



Consider a dataset $\mathcal{D}$ with $n$ entries where each entry is a random tuple $(S,X,Y)\in \mathcal{S}\times\mathcal{X}\times\mathcal{Y}$ where $S$, $X$, and $Y$ are sensitive, non-sensitive, and target (non-sensitive) features, respectively, and $\hat{Y}\in \mathcal{Y}$ is a predictor of $Y$.
{Note that $S$ and $Y$ can be a collection of features or labels (e.g., $S$ can be gender, race, sexual orientation, or a combination of these, while $Y$ could be age, facial expression, etc.); for ease of writing, we use the term variable to denote both single and multiple features/labels}. 
Instances of $X$, $S$, and $Y$ are denoted by $x$, $s$ and $y$, respectively. The entries $(X,S,Y)$ of $\mathcal{D}$ are independent and identically distributed (i.i.d.)
according to $P(X,S,Y)$. 

Recent results on algorithmic fairness guarantee that, for a specific target $Y$, the prediction of a machine learning model is accurate with respect to (\textit{w.r.t.}) $Y$ but unbiased \textit{w.r.t.} the sensitive $S$. While more than two dozen measures for fairness have been proposed, two oft-used fairness measures are demographic parity and equalized odds (and variants thereof). 
Demographic parity (DemP) ensures complete independence between the prediction of the target $\hat{Y}$ and the sensitive $S$; this notion of fairness favors utility the least, especially when $Y$ and $S$ are correlated \cite{hardt2016equality}. Equalized odds (EO) enforces this independence conditioned on $Y$, thereby ensuring equal rates for true and false positives (when $Y$ is binary) for all demographics. We now define DemP and EO formally (for binary $S$ and $Y$ as originally introduced). These definitions can be generalized to the non-binary setting and we do so in the sequel for fair representations. 



\begin{definition}[\cite{hardt2016equality}]\label{Def:ThreeFairnessMeasures}
A predictor $f(S,X)=\hat{Y}$ satisfies 
\begin{itemize}[leftmargin=*]
    \item demographic parity (DemP) w.r.t. $S$, if $\hat{Y} \perp S$, i.e., 
\thickmuskip=2mu
\begin{align}
\label{eq:demographic_parity}
    \emph{Pr}(\hat{Y}=1|S=1)=\emph{Pr}(\hat{Y}=1|S=0)
\end{align}
    \item equalized odds (EO) \textit{w.r.t.} $(S,Y)$, if $\hat{Y} \perp S|Y$, for $y\in \{0, 1\}$:
\begin{align}
\label{eq:equalied_odds}
    \emph{Pr}(\hat{Y}=1|S=1, Y=y)=\emph{Pr}(\hat{Y}=1|S=0, Y=y).
\end{align}
\end{itemize}
\end{definition}

In the following section, we present our FUR framework which includes definitions for both demographically fair and censored representations.
\vspace{-0.1in}
\section{FURs via Generative Adversarial Models}
\label{sec:model}


Formally, the FUR model consists of two components, an encoder and an adversary, as shown in Fig. \ref{fig:gap}. The goal of the encoder $g:\mathcal{X}\times\mathcal{S} \to \mathcal{X}_r$ is to actively eliminate the dependence between $S$ and $X$ while that of the adversary $h:\mathcal{X}_r \to \mathcal{S}$ is to infer $S$. In general, $g(X,S)$ is a randomized mapping that outputs a representation $X_r = g(X,S)$. Note that $S$ {may not always} be available to the curator; however, it will always affect the design of $g$ via the adversarial training process. For brevity, we henceforth write $g(\cdot)$ to include both possibilities (just $X$ or $(X,S)$ as inputs). 
On the other hand, the role of the adversary is captured via $h(X_r)$, which is the adversarial decision rule 
in inferring the sensitive variable $S$ as $\hat{S}=h(X_r=g(\cdot))$ from the representation $g(\cdot)$. In general, the hypothesis $h$ can be a \textit{hard decision rule} under which $h(g(\cdot))$ is a direct estimate of $S$ or a \textit{soft decision rule} under which $h(g(\cdot)) = P_{h}(\cdot|g(\cdot))$ is a distribution over $\mathcal{S}$.

To quantify the adversary's performance, we use a loss function $\ell(h(g(X=x,S=s)),S=s)$ defined for every pair $(x,s)$. Thus, the adversary's {expected loss} \textit{w.r.t.} $X$ and $S$ is $L(h, g)\triangleq \mathbb{E}[\ell(h(g(\cdot)),S)]$, where the expectation is taken over $P(X,S)$ and the randomness in $g$ and $h$.
To ensure utility, we introduce a constraint on the fidelity of $X_r$ via a distortion function 
$d(x_r,x)$, which measures the goodness of $X_r = x_r$ \emph{w.r.t.} $X = x$. Ensuring statistical utility, in turn, requires constraining the average distortion $\mathbb{E}[d(g(\cdot),X)]$, where the expectation is taken over $P(X,S)$ and the randomness in $g$.

\vspace{-0.1in}
\subsection{FUR: Framework and Theoretical Results}
To publish a fair representation $X_r$, the data curator wishes to learn an encoder $g$ that guarantees censoring (i.e., it is difficult for the adversary to learn $S$ from $X_r$), and therefore, fair $X_r$ under DemP, as well as utility ($g$ guarantees bounded distortion of $X$). In contrast, for a fixed $g$, the adversary would like to find a (potentially randomized) function $h$ that {minimizes its expected loss, or equivalently maximizes the negative expected loss}. This leads to a constrained minimax game between the encoder and the adversary given by
\begin{equation}
\label{eq:generalopt}
\min_{g(\cdot)}\max_{h(\cdot)} -\mathbb{E}[\ell(h(g(\cdot)),S)], \: \textrm{s.t.} \: \mathbb{E}[d(g(\cdot),X)]\le D.
\end{equation}
where $D\ge0$ determines the distortion constraint on $X_r$. The optimization in \eqref{eq:generalopt} highlights that the input to $g$ depends on whether the curator has access to both $(X,S)$ or just $X$. Having access to both $(X,S)$
in general will yield a better decorrelator (e.g., see Section \ref{sec:uci_adult} for the UCI dataset).
Finally, without the constraint in \eqref{eq:generalopt}, the optimal $X_r=g(\cdot)\perp S$. One can approximate this in practice via arbitrarily large distortions as we show in Proposition \ref{Thm:GAP_censoring}; as a setup to these results, we first 
define censoring and fairness for representations. Our censoring definition clarifies the representation that 
best limits an adversary from inferring $S$. We then define DemP for FRs; we combine the two definitions to show how and when adversarial learning can help ensure demographic parity.  

\begin{definition}[Censored Representations]\label{Def:Censored_representation}
A representation $X_r$ of $X$ is censored \textit{w.r.t.} the sensitive features $S$ 
against a learning adversary $h(\cdot)$, whose performance is evaluated via a loss function $\ell(h(X_r),S)$, if for an optimal adversarial strategy $h^*=\argmin_{h}\mathbb{E}[\ell(h(X_r),S)]$,
\begin{align} 
\label{eq:Censored_representation}
    \mathbb{E}[\ell(h^*(g(\cdot)),S)] \le \mathbb{E}[\ell(h^*(X_r),S)],
\end{align}
where $g(\cdot)$ is any (randomized) function of $X$ (or $(X,S)$) and the expectation is over $h$, $g$, $X$, and $S$.
\end{definition}
The above definition suggests that the best censored representation $X_r$ is the least informative about $S$ to an adversary whose inferential action is captured by a loss function $\ell(\cdot,\cdot)$, i.e., the average loss is the worst for $X_r$ than for any other arbitrary function $g(\cdot)$. While the comparison in \eqref{eq:Censored_representation} is w.r.t. the best $h^*(X_r)$ for $X_r$, choosing the optimal $h(\cdot)$ for any $g(\cdot)$ will only serve as a lower bound to the left side of \eqref{eq:Censored_representation}.

We now define DemP for representations; we then prove that a demographically fair representation $X_r$ guarantees that any downstream algorithm using $X_r$ satisfies DemP \textit{w.r.t.} $S$. {In the following, we assume that $X$, and therefore, $X_r$ are discrete random variables with arbitrarily large alphabets; however, the definition below can be extended to continuous-valued $X$ and $X_r$ by considering all Borel subsets and an appropriately defined measure on the space.} 
\begin{definition}[Demographically Fair Representations]\label{Def:DemP_representation}
For $(X,S)\in\mathcal{X}\times\mathcal{S}$, a representation $X_r=g(X,S)\in\mathcal{X}_r$ satisfies demographic parity \textit{w.r.t.} $S$ if for any $x_r\in \mathcal{X}_r$ and $s,s'\in \mathcal{S}$
\begin{align}
\label{eq:DemP_representation}
    \emph{Pr}(X_r=x_r|S=s)=\emph{Pr}(X_r=x_r|S=s')
\end{align}
where $g:\mathcal{X}\times\mathcal{S}\rightarrow\mathcal{X}_r$ is any (possibly randomized) function.
\end{definition}


\begin{theorem}[Fair Learning via Fair Representation]\label{thm:DemP_broadcasting}
If $X_r=g(X,S)$ satisfies demographic parity \textit{w.r.t.} $S$, then any algorithm $f:\mathcal{X}_r \to \mathcal{Y}$ satisfies demographic parity \textit{w.r.t.} $S$.
\end{theorem}
{
\if \extended 0
The proof of Theorem \ref{thm:DemP_broadcasting} follows from a direct application of the data-processing inequality for mutual information since $(X,S)-X_r-Y$ form a Markov chain; details can be found in \cite{CFUR-LS}. 
\fi}
\if \extended 1 The proof of Theorem \ref{thm:DemP_broadcasting} follows from the data-processing inequality of mutual information as detailed in Appendix \ref{Proof:thm:DemP_broadcasting}. \fi

\begin{remark} Note that equalized odds in Def. \ref{Def:ThreeFairnessMeasures} explicitly involves a downstream task, and therefore, the design of an EO fair $X_r$ needs to include a predictor explicitly. In contrast to the FUR setting considered here, such targeted representations and the ensuing fair predictors provide guarantees only for specific target $Y$. In this limited context, however, one can still define an $X_r$ as ensuring EO \textit{w.r.t.} to $(S,Y)$ if $\hat{Y}(X_r)\perp S|Y$.
\end{remark}

One simple approach to obtain a fair/censored representation $X_r$ is by choosing $X_r=N$ where $N \perp (X,S)$. 
However, such an $X_r$ has no utility (quantified, for example, via downstream task accuracy). The design of $X_r$ has to ensure utility, and thus, there is a tradeoff between guaranteeing fairness/censoring and assuring a desired level of utility. We now quantify such tradeoffs using FUR framework. 

\begin{theorem}\label{Thm:GAP_censoring}
For sufficiently large distortion bound $D$, \eqref{eq:generalopt} yields a universal representation $X_r$ censored \textit{w.r.t.} to $S$.
\end{theorem}
The proof follows by observing that for sufficiently large $D$, $X_r$ can be arbitrarily noisy, reducing \eqref{eq:generalopt} to an unconstrained optimization.
For this $X_r$ with $h^*=\argmin_{h}\mathbb{E}[\ell(h(X_r),S)]$, 
\begin{align}
    \mathbb{E}[\ell(h^*(X_r),S)] &=-\min_{g(\cdot)} \max_{h(\cdot)}-\mathbb{E}[\ell(h(g(\cdot)),S)]\\
     &\geq  \mathbb{E}[\ell(h^*(g(\cdot)),S)],
\end{align}
thus satisfying Definition \ref{Def:Censored_representation}. 



The FUR framework in \eqref{eq:generalopt} places no restrictions on the adversary. Indeed, different loss functions and decision rules lead to different adversarial models (see Table~\ref{table:gaploss}). This versatility to a large class of (inferring) adversarial models is captured by the last entry in Table~\ref{table:gaploss} by using the recently introduced tunable \textit{$\alpha$-loss} \cite{Liao2019TIT,sypherd2021tunable}, defined for $\alpha \in (0,1) \cup (1,\infty)$ as :
\begin{align}
\label{eq:alpha_loss}
    \ell_\alpha (h(g(\cdot)),s) =\frac{\alpha}{\alpha  - 1} \left( 1- P_h(s|g(\cdot))^{ \frac{\alpha-1}{\alpha}}\right)
\end{align} 
with continuous extensions at $\alpha=1$ and $\alpha=\infty$ (the loss simplifies to a constant for $\alpha=0$). Note that the loss in \eqref{eq:alpha_loss} operates on a soft decision $P_h(\cdot|\cdot)$, the output of the adversary $h$. 
By tuning $\alpha \in [0,\infty]$, $\alpha$-loss captures a variety of information-theoretic adversaries as listed in Table~\ref{table:gaploss} (see also \cite{Liao2019TIT,sypherd2021tunable}):\\ 
(i) a hard-decision adversary for $\alpha=\infty$ captured by $\ell_{\infty}(h(g(\cdot)),s) = 1-\text{Pr}[{h(g(\cdot)) = s}]$\footnote{For $\alpha=\infty$, $\alpha$-loss reduces to probability of error, for which the optimal rule that minimizes the expected loss is the \textit{maximal a posteriori} (MAP) estimation, a hard decision. The same rule results when minimizing the 0-1 loss which is given by $\mathbb{I}({h(g(\cdot))\neq s})$ where $\mathbb{I}$ is the indicator function.}, and \\ 
(ii) a soft-decision adversary for $\alpha=1$ via the oft-used log-loss  $\ell_1(h(g(\cdot)),s) = -\log {P_h(s|g(\cdot))}$ ({this follows directly by applying L'H\^opital's rule)}.\\
(iii) Values of $\alpha>1$ allow interpolating between the NP-hard to implement MAP rule ($\alpha=\infty$) and log-loss ($\alpha=1$) and allow some robustness to noisy data \cite{sypherd2021tunable}. On the other hand, choosing $\alpha<1$ leads to more convex losses than log-loss ($\alpha=1/2$ yields a soft exponential loss used in boosting algorithms) that are more sensitive to outliers \cite{sypherd2021tunable}. 

For any encoder $g(\cdot)$, the following proposition (see also the last row of Table~\ref{table:gaploss}) summarizes the optimal $P^*_h(s|g(\cdot))$ under $\alpha$-loss.




\begin{proposition}
\label{thm:alpha}
For a fixed $g$, under $\alpha$-loss, the optimal adversary decision rule that minimizes the expected loss is a `$\alpha$-tilted' conditional distribution
$P^*_h(s|g(\cdot)) = \frac{P(s|g(\cdot))^\alpha}{\sum_{s\in\mathcal{S}} P(s|g(\cdot))^\alpha}.$ For $\alpha=1$ and $\alpha=\infty$, we obtain the optimal strategies for log-loss and $0$-$1$ loss, respectively, as the true conditional distribution and the maximal \textit{a posteriori} (MAP) estimator. Then, \eqref{eq:generalopt} reduces to
$\min_{g(\cdot)} -H^{\text{A}}_{\alpha}(S|g(\cdot))$, 
where $H^{\text{A}}_{\alpha}(\cdot|\cdot)$\footnote{$H^{\text{A}}_{\alpha}(U|V)\triangleq 1/(1-\alpha)\log(\sum_{u,v}P_{U,V}^\alpha(u,v))$} is the Arimoto conditional entropy. 
\end{proposition}

Proposition \ref{thm:alpha} states that if the adversary uses $\ell_\infty$, 
\eqref{eq:generalopt} simplifies to $\min_{g(\cdot)} P(g(\cdot))\max_{s\in S} P(s|g(\cdot))-1$, i.e., we choose the most likely $s$ for every $g(\cdot)$ (the MAP rule)~\cite{asoodeh2018estimation}.


On the other hand, if the adversary uses log-loss, 
\eqref{eq:generalopt} simplifies to $\min_{g(\cdot)}I(g(\cdot);S)$ for any prior on $S$, 
where $I(g(\cdot);S)$ is the mutual information (MI) between $g(\cdot)$ and $S$. More generally, using $\alpha$-loss in \eqref{eq:alpha_loss}, the optimal $P_h^*$ in Proposition \ref{thm:alpha} simplifies the objective in \eqref{eq:generalopt} to $\min_{g(\cdot)}I_\alpha^\text{A}(g(\cdot);S)$, where $I_\alpha^\text{A}(g(\cdot);S)$ is the Arimoto MI\footnote{$I_\alpha^\text{A}(g(\cdot);S)=H_\alpha(S)-H_\alpha^\text{A}(S|g(\cdot))$ where $H_\alpha(S)\triangleq H_\alpha^\text{A}(S)$ is the R\'enyi entropy of order $\alpha$ and is also the unconditioned $\alpha$-Arimoto entropy.} of order $\alpha$. These MIs can be effective proxies for guaranteeing DemP FURs, since they are minimized only when $S \perp g(\cdot)$, thus leading to the following theorem.

\begin{table*}[]
\small
\centering
\begin{tabular}{|c|c|c|c|}
\hline
              & \begin{tabular}[c]{@{}l@{}}Loss function $\ell(h(g(\cdot)),s) $\end{tabular}                                                                              & \begin{tabular}[c]{@{}l@{}}Optimal adversarial strategy $h^*$\end{tabular} & Adversary type                                                                           \\ \hline
Squared loss  & $ (h(g(\cdot)) - S)^2$                                                                                                                                       & $\mathbb{E}[S|g(\cdot)]$                                                          & MMSE adversary                                                                           \\ \hline
0-1 loss      & $
\left\{
\begin{array}{ll}
0  & \mbox{if } h(g(\cdot))=S \\
1  & \mbox{otherwise }
\end{array}
\right.$ & $\argmax\limits_{s \in \mathcal{S}} P(s|g(\cdot))$                                & MAP adversary                                                                            \\ \hline
Log-loss      & -$\log {P_h(s|g(\cdot))}$                                                                                                                             & $P(s|g(\cdot))$                                                                   & \begin{tabular}[c]{@{}l@{}}Belief refining adversary \end{tabular} \\ \hline
$\alpha$-loss & $\frac{\alpha}{\alpha  - 1} \left( 1- P_h(s|g(\cdot))^{1 - \frac{1}{\alpha}}\right)$                                                                         & $\frac{P(s|g(\cdot))^\alpha}{\sum_{s\in\mathcal{S}} P(s|g(\cdot))^\alpha}$    & \begin{tabular}[c]{@{}l@{}}Generalized belief \\ refining adversary\end{tabular}         \\ \hline
\end{tabular}
\caption{The adversaries captured by the FUR framework by using a variety of loss functions.}
\label{table:gaploss}
\end{table*}

\begin{theorem}
\label{thm:DP}
{Under $\alpha$-loss, for all $\alpha$, \eqref{eq:generalopt} enforces fairness subject to a distortion constraint. As the distortion increases, the ensuing fairness guarantee approaches ideal DemP. }
\end{theorem}

{Proposition \ref{thm:alpha} 
was proved in \cite{huang2017context}; Theorem \ref{thm:DP} involves similar arguments to Theorem \ref{Thm:GAP_censoring}. \if \extended 0
All proofs can be found in \cite{CFUR-LS}. \fi
\if \extended 1 We combine the proofs for Proposition \ref{thm:alpha} 
and Theorem \ref{thm:DP} in Appendix \ref{proof:table:gaploss}. \fi }
{Many notions of fairness (cf. Definition \ref{Def:ThreeFairnessMeasures}) require computing conditional probabilities for every sample $x$ to ensure independence, and thus, are not easy to optimize in a data-driven fashion. The FUR framework via loss functions captures mutual information-like surrogates for such independence conditions; to this end, Theorem \ref{thm:DP} justifies using $\alpha$-loss (and thus, log-loss too) as a proxy for enforcing fairness}. 
{We remark that mutual information (MI) is a common surrogate fairness measure for demographic/statistical parity \cite{song19a-fair,calmon2017optimized,Calmon2018data}. In  \cite{song19a-fair}, the authors use MI for both fairness and the distortion measure leading to an (non-convex) information bottleneck problem; we recover this formulation by choosing $d(x,x_r)=-\log p(x_r|x)$. Thus the FUR framework is more general and allows choosing application-dependent meaningful fidelity measures (for example, different measures of representation similarity are used in natural language processing and healthcare data).}  
{Finally, we remark that the optimal adversarial strategy in Proposition \ref{thm:alpha} requires estimating a posterior; as described in Section \ref{subsec:data-driven}, in practice, one can use deep learning models for $h$ and $g$ to do so.}


{
A predominant approach in the literature in the context of fair representations is to explicitly include the intended classification/prediction task, i.e., design representations that guarantee DemP for the specific task \cite{Madras_2018,edwards_Storkey2016,zhang2018mitigating}. In fact, the FUR formulation in \eqref{eq:generalopt} can be extended to include this case by adding an additional term in the objective function to ensure high accuracy in learning $Y$. The resulting minimax game is given by
\begin{subequations}
\label{eq:generalopt-oneApp}
\begin{align}
\min_{\tilde{g}(\cdot),f(\cdot)}\max_{h(\cdot)} -&\mathbb{E}[\ell(h(\tilde{g}(\cdot)),S)]+\lambda \mathbb{E}[\ell'(f(\tilde{g}(\cdot)),Y)],\\
\textrm{s.t.}  \quad & \mathbb{E}[d(\tilde{g}(\cdot),X)]\le D, \label{eq:generalopt-oneApp-b}
\end{align}
\end{subequations}
where $f(\cdot)$ is a classifier for a target $Y$, $\lambda> 0$, and $\tilde{g}(\cdot)$\footnote{In general, $\tilde{g}(\cdot)$ can be a function of both $X$ and $S$; the dependence on $S$ is implicit when $S$ is not directly available.}  
and $h(\cdot)$ are the encoder and the adversarial classifier, respectively, as in \eqref{eq:generalopt}.  
Note that the loss functions $\ell(\cdot)$ and $\ell'(\cdot)$ can be different.
The setup in \eqref{eq:generalopt-oneApp} involves an additional term ensuring fair classification and is, thus, a more constrained optimization than the FUR framework; in fact, we recover the FUR setup with $\lambda=0$. However, even while generating intermediate representations $g(\cdot)$, \eqref{eq:generalopt-oneApp} is primarily intended to design \textit{fair classifiers}, and therefore, requires knowing the intended tasks on $Y$. In contrast, our FUR framework allows generating DemP-guaranteeing fair $X_r$ that in turn guarantee DemP fairness to all downstream tasks on any subset of $Y$.  
}



One can also design fair classifiers directly without intermediate representations by setting $\tilde{g}(\cdot)\triangleq \hat{Y}$; such classifiers $\tilde{g}$ can be designed with either DemP or EO guarantees. 
We first consider the more general problem of designing EO-fair predictors/classifiers $\tilde{g}(\cdot)$ for the target $Y$ and show that our FUR framework subsumes this problem (and therefore, that of generating DemP predictors/classifiers). Let $h$ be the adversary decision rule to infer $S$ as $\hat{S}=h(\tilde{g}(\cdot)|Y)$ for every choice of $Y$ via the soft predictor $\tilde{g}(\cdot)=P_{\hat{Y}|\cdot}$. 
Then, analogous to \eqref{eq:generalopt}, the design of a fair predictor/classifier can be formulated as
\begin{equation}
\label{eq:Opt-fairClassifier-EO}
\min_{\tilde{g}(\cdot)}\max_{h(\cdot)}  -\mathbb{E}\big[\ell\big(h(\tilde{g}(\cdot)|Y),S\big)\big], \:
\textrm{s.t.} \: \mathbb{E}[\ell\big(\tilde{g}(\cdot),Y\big)]\le \epsilon,
\end{equation}
{where the expectation now includes $Y$ too}. 

\begin{theorem}
\label{prop:EO}
Under $\alpha$-loss, 
the formulation in \eqref{eq:Opt-fairClassifier-EO} enforces EO fairness subject to a performance constraint $\epsilon$. As $\epsilon$ increases, the ensuing fairness guarantee approaches the ideal EO guarantees achievable by $\tilde{g}$ w.r.t. $S$ and $Y$. 
\end{theorem}
{
\if \extended 0
The proof of Theorem \ref{prop:EO} is similar to that for Theorem \ref{Thm:GAP_censoring} and can be found in \cite{CFUR-LS}. \fi}
\if \extended 1 The proof of Theorem \ref{prop:EO} is in Appendix \ref{Proof:prop:EO}. \fi 
Note that the formulation in \eqref{eq:Opt-fairClassifier-EO} also holds for generating a fair predictor/classifier satisfying DemP in Definition \ref{Def:ThreeFairnessMeasures}. In contrast to EO where the adversary needs both $\tilde{g}(\cdot)$ and $Y$ as inputs, for DemP, only $\tilde{g}(\cdot)$ is input to the adversary. 

The adversarial models and the resulting game-theoeretic solutions in Table \ref{table:gaploss} highlight the formal guarantees of the FUR framework. Recently, Sypherd \emph{et al.} have demonstrated the robustness of training deep learning models with $\alpha$-loss to both noise and class imbalances \cite{sypherd2021tunable}, thus promising to be applicable for learning FURs with GANs. The rest of the sequel focuses on data-driven GANs with $\alpha=1$, i.e., log-loss, to highlight the value of FURs in guaranteeing fairness for multiple downstream tasks relative to the state-of-the-art fair classifiers. Future work will include enhancing such results to include tuning over $\alpha$.  

\vspace{-0.15in}
\subsection{Data-driven FUR}\label{subsec:data-driven}
Thus far, we have focused on a setting where the curator has access to the statistics $P(X,S)$ thereby solving the constrained minimax optimization problem in \eqref{eq:generalopt} (game-theoretic version of the FUR formulation) to obtain a $g$ that performs best against a chosen adversary. In practice, $P(X,S)$ is impossible to compute. To this end, we propose a data-driven version of the FUR formulation that allows the data holder to learn a generative decorrelator from an $n$-sample dataset $\mathcal{D} = \{(x_{(i)},s_{(i)})\}_{i = 1}^{n}$ via a generative model $g(X;\theta_{p})$ that is parameterized by $\theta_{p}$. This generative model takes $X$ (or $(X,S)$) as input and outputs $X_r$. In the training phase, the data holder learns the optimal parameters $\theta_p$ by competing against a \textit{computational adversary}: a classifier modeled by a neural network $h(g(X;\theta_{p}); \theta_a)$ that is parameterized by $\theta_{a}$. In the evaluation phase, the performance of the learned decorrelation scheme can be tested under a strong adversary that is computationally unbounded and has access to dataset statistics. We follow this procedure in the next section.


While, in theory, the functions $h$ and $g$ can be arbitrary, in practice, they are best approximated by a well-chosen rich hypothesis class. Fig.~\ref{fig:gap} illustrates a FUR model in which $h$ and $g$ 
are both modeled as deep neural networks (DNNs). For a fixed $h$ and $g$, binary $S$ and $\ell=\ell_1$ (log-loss for $\alpha=1$), the adversary's {\textit{empirical loss}} using cross entropy is given by
\begin{equation}
\label{eq:gap-CE-datadriven}
\begin{aligned}
    L_{n}(\theta_p,\theta_a)&=  -\frac{1}{n}\sum\limits_{i=1}^{n}s_{(i)}\log h(g(x_{(i)}; \theta_{p});\theta_{a}) \\
    &\quad +(1-s_{(i)})\log(1-h(g(x_{(i)}; \theta_{p});\theta_{a})).
\end{aligned}
\end{equation}
The optimal model parameters $(\theta_p,\theta_a)$ are then solutions of
\begin{equation}
	\label{eq:learnedprivatizer}
	\min_{\theta_{p}}\max\limits_{\theta_a} -L_{{n}}(\theta_p,\theta_a), \:
	\textrm{s.t.} \:
	\frac{1}{n}\sum\limits_{i=1}^{n}d(g({x}_{(i)};\theta_{p}), {x}_{(i)})\le D.
\end{equation}

The minimax optimization in \eqref{eq:learnedprivatizer} is a two-player non-cooperative game between the generative decorrelator and the adversary with strategies $\theta_{p}$ and $\theta_{a}$, respectively. In practice, for chosen hypothesis classes for $g$ and $h$ (e.g., DNN architectures), we can learn the equilibrium of the game using an iterative algorithm as follows. 
(i) For a fixed $\theta_p$, maximize the negative of the adversary's loss to compute the parameters of $h$. (ii) Then, minimize the decorrelator's loss (negative adversary loss) to compute $\theta_p$ for a fixed $h$. 

It is crucial to note that the \emph{hard} distortion constraint in \eqref{eq:learnedprivatizer} makes our minimax problem different from what has been extensively studied in the literature. 
To incorporate the distortion constraint, we use the \textit{penalty method} \cite{lillo1993solving} to replace the constrained optimization problem by adding a penalty to the objective function. This is done via a penalty parameter $\rho_t$ that captures a measure of violation of the constraint at the $t^{\text{th}}$ iteration. The constrained optimization problem of $g$ is then approximated by a \emph{series of unconstrained optimization problems} with an objective
\begin{equation}
    -L_n(\theta_{p},\theta_a)+\rho_t(\max\{0,\frac{1}{n}\sum_{i=1}^{n}d(g({x}_{(i)};\theta_{p}), {x}_{(i)})-D\})^2,
    \label{eq:penaltymethodsupp}
\end{equation}
where the penalty coefficient $\rho_t$ decreases with the number of iterations $t$. {
We note that both the augmented Lagrangian and the penalty methods have similar performance in practice; we chose the penalty method but our results can also be obtained with the augmented Lagrangian method~\cite{eckstein2012augmented}.}
{We provide detailed steps of the algorithm and the parameters for the penalty method in Appendix~\ref{sec:alternateminimax}; we also clarify our methodology for choosing both $\rho_t$ and the learning rate $\eta_t$ there}. Finally, we note that one can easily generalize \eqref{eq:gap-CE-datadriven} to the multi-class setting (non-binary $S$) using the softmax function; one can also generalize \eqref{eq:gap-CE-datadriven} using $\alpha$-loss.

%
{In the following sections, we detail our results for synthetic multi-dimensional Gaussian mixture data and four publicly available datasets: UCI Adult, UTKFace, GENKI, and HAR.}
All code is available via GitHub at \cite{CFUR-LS}. 

\vspace{-0.08in}
\section{FUR for Gaussian Mixture Models}
\label{sec:gmm}
In this section, we focus on a setting where $S \in \{0, 1\}$ and ${X}$ is an $m$-dimensional Gaussian mixture random vector whose mean is dependent on $S$. Let $P(S =1) = q$. Let ${X}|S =0 \sim \mathcal{N}(-{\mu}, \Sigma)$ and $ {X}|S =1 \sim \mathcal{N}({\mu}, \Sigma)$, where  ${\mu}=(\mu_1,...,\mu_m)$. We assume that ${X}|S=0$ and ${X}|S=1$ have the same covariance $\Sigma$.
\vspace{-.15in}
\subsection{Game-Theoretical Approach}
We consider a MAP adversary that has access to $P({X}, S)$ and $g$. The goal is to censor ${X}$ in a way that minimizes the adversary's probability of correctly inferring $S$ from $X_r$. In order to have a tractable model for the encoder, we mainly focus on affine representations $X_r=g({X})={X}+ {Z}+{\beta}$, where for tractability reasons, we choose ${Z}$ as a zero-mean multi-dimensional Gaussian random vector independent of $X$. 
This linear representation enables controlling both the mean and covariance of $X_r$.
To quantify utility of the privatized data, we use the $\ell_2$ distance between ${X}$ and $X_r$ as a distortion measure to obtain a constraint $\mathbb{E}_{{X},X_r}\lVert {X}-X_r\rVert^2\le D$.



We assume that ${\beta}=(\beta_1,...,\beta_m)$ is a constant parameter vector and ${Z}\sim\mathcal{N}(0,\Sigma_p)$. Building on the analysis in \cite{gallager2013stochastic}, for the standard Gaussian $Q(\cdot)$ function, we can derive the adversary's detection probability $P^{(G)}_{d}$ as
\begin{equation}
\begin{aligned}
\label{eq:gaussianscheme0}
P^{(G)}_{d}&= q Q\left(-  {\gamma \over 2}  + {1 \over \gamma } \ln\left( { 1  - q  \over q} \right) \right) \\
& \quad + (1 - q) Q\left(-  {\gamma \over 2}  - {1 \over \gamma } \ln\left( { 1  - q  \over q} \right) \right),
\end{aligned}
\end{equation}
where 
$\gamma=\sqrt{(2{\mu})^T(\Sigma+\Sigma_p)^{-1}2{\mu}}$. From the constraint, we have 
$\mathbb{E}_{{X},X_r}\lVert {X}-X_r\rVert^2= 
\lVert{\beta}\rVert^2+tr(\Sigma_p) \leq D$. {The mixture Gaussian classification problem, especially for the same covariance for both classes $S=0$ and $S=1$, is a tractable problem that has been studied in a variety of settings including communication systems \cite{gallager2013stochastic} and machine learning \cite{hastie01statisticallearning}, to name a few. The result in \eqref{eq:gaussianscheme0} builds directly on \cite{gallager2013stochastic}, and so, for reasons of brevity, we leave it out.} 
The following theorem summarizes the optimal noising strategy when ${X}|S$ and ${Z}$ are multi-dimensional {i.i.d.} Gaussians. 


\begin{theorem}
\label{thm:PDI}
Consider the representation given by $g({X})={X}+ {Z}+{\beta}$, where ${X|S}$ and ${Z}$ are Gaussian random vectors with diagonal covariance matrices $\Sigma$ and $\Sigma_p$, respectively, and ${X|S}\perp {Z}$. Let $\{\sigma^2_1,...,\sigma^2_m\}$ and $\{\sigma^2_{p_{1}},...,\sigma^2_{p_{m}}\}$ be the diagonal entries of $\Sigma$ and $\Sigma_p$, respectively. The parameters of the minimax optimal censoring mechanism $g^*$ are
$${\beta_i}^*={0}, \quad{\sigma^*_{p_{i}}}^2=\left(\frac{\lvert\mu_i\rvert}{\sqrt{\lambda^*_0}}-\sigma_i^2\right)^{+}, \: \forall i=\{1,2,...,m\},$$
where $\lambda^*_0$, the dual variable enforcing the distortion constraint in \eqref{eq:generalopt}, is chosen such that $\sum_{i=1}^{m} {\sigma^*_{p_{i}}}^2=D$.
For this optimal mechanism, the accuracy of the MAP adversary is given by \eqref{eq:gaussianscheme0} with
$\gamma=2\sqrt{\sum_{i=1}^{m} {\mu_i^2}/{(\sigma_i^2+{\sigma^*_{p_{i}}}^2)}}.$
\end{theorem}
The proof of Theorem \ref{thm:PDI} is in Appendix \ref{proof:PDI}. We observe that when $\sigma^2_i>{\lvert\mu_i\rvert}/{\sqrt{\lambda^*_0}}$, no noise is added to the data on this dimension due to the high variance. In contrast, when $\sigma^2_i < {\lvert\mu_i\rvert}/{\sqrt{\lambda^*_0}}$, the variance of the noise added to this dimension is proportional to $\lvert\mu_i\rvert$; this is intuitive since a large $\lvert\mu_i\rvert$ indicates the two conditionally Gaussian distributions are further away on this dimension, and thus, require more noise to reduce the MAP adversary's inference accuracy. 

{We note that we could have considered a more general non-affine model for the encoder. Since synthetic datasets provide a verifiable way to formally evaluate the FUR framework, we chose a simpler affine generative model that is tractable and yields closed-form information-theoretic results, i.e., we can derive the best adversarial decoder. This in turn is helpful in comparing the data-driven approach with the game-theoretic optimal solution on these canonical data models as a much-needed sanity check. Finally, the analysis here can be generalized to correlated Gaussian distributions for each sensitive group and one expects a similar behavior as the features can be whitened when the covariance is the same for both classes.}

\vspace{-0.13in}
\subsection{Data-driven Approach}
To learn the data-driven representation $X_r=g({X})={X}+ {Z}+{\beta}$ using our FUR framework, we assume that $g$ only has access to the dataset $\mathcal{D}$ with $n$ data samples (not $P(X,S)$). Computing the optimal $g^*$ is then a learning problem. In the training phase, we learn the parameters $\theta_p$ of $g$ by competing against a computational adversary $h(g(\theta_p);\theta_a)$ modeled by a multi-layer neural network. When convergence is reached, we evaluate the performance of the learned mechanism by comparing with the one obtained from the game-theoretic approach. To quantify the performance of the learned $X_r$, we compute the accuracy of inferring $S$ under a strong MAP adversary that has access to both the joint distribution of $({X}, S)$ and the censoring mechanism.

Since the sensitive variable $S$ is binary, we measure the training loss of the adversary network using the empirical log-loss function in \eqref{eq:gap-CE-datadriven}. 
We model the encoder using a two-layer neural network with parameters $\theta_p=\{\beta_1, ..., \beta_m, \sigma_{p_1},..., \sigma_{p_m}\}$, where $\beta_k$ and $\sigma_{p_k}$ represent the mean and standard deviation for each dimension $k\in\{1,...,m\}$, respectively. The random noise ${Z}$ is drawn from a $m$-dimensional independent zero-mean standard Gaussian distribution such that {$\{\sigma_{p_k}Z_k\}_{k=1}^{m}$} jointly have a covariance $\Sigma_p=\diag(\sigma^2_{p_{1}},...,\sigma^2_{p_{m}})$. Thus,  $\hat{X}_k=X_k+\beta_k+\sigma_{p_k}Z_k$. The adversary 
is modeled by a three-layer neural network classifier with leaky ReLU activations. Finally, as detailed in Algorithm \ref{alg:euclid}, we use the penalty method to ensure the distortion constraint.

\vspace{-0.15in}
\subsection{Illustration of Results}
\label{sec:GMM-results}

We generate two synthetic datasets to illustrate our results; each dataset has $20K$ training samples and $2K$ test samples. Each dataset is generated by sampling from an independent multi-dimensional Gaussian mixture model.
The two datasets correspond to two distinct values for the prior $P(S=1)$ as $0.75$ and $0.5$. Both encoder and adversary are trained via Tensorflow \cite{abadi2016tensorflow} using the Adam optimizer \cite{kingma2017adam} with a learning rate of $0.005$ and a minibatch size of $1000$. 

\begin{figure}[!ht]
\vspace{-0.12in}
	\centering
	\begin{subfigure}[t]{0.4\textwidth}
        \includegraphics[width=1\columnwidth,trim=0 8 0 11,clip]{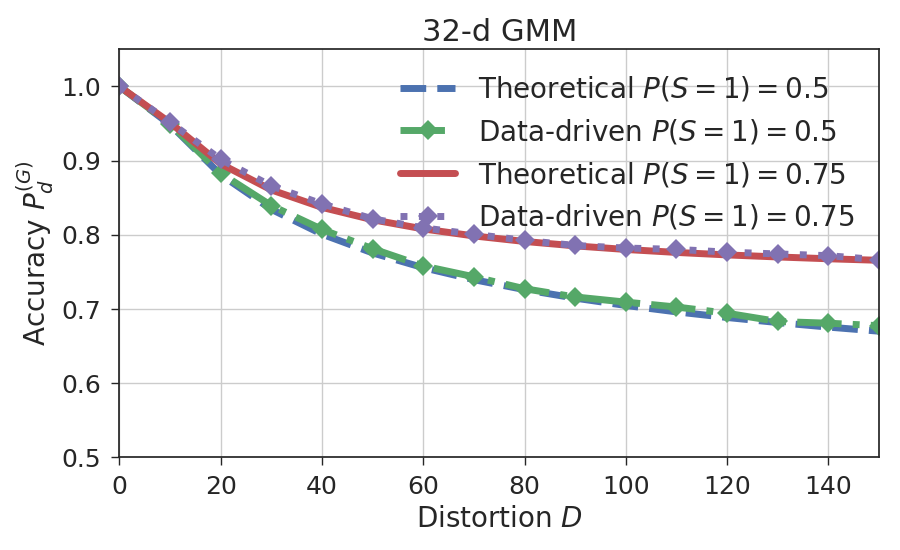}
		\caption{Sensitive variable classification accuracy}
	\end{subfigure}
	\begin{subfigure}[t]{0.4\textwidth}
        \includegraphics[width=1\columnwidth,trim=0 8 0 11,clip]{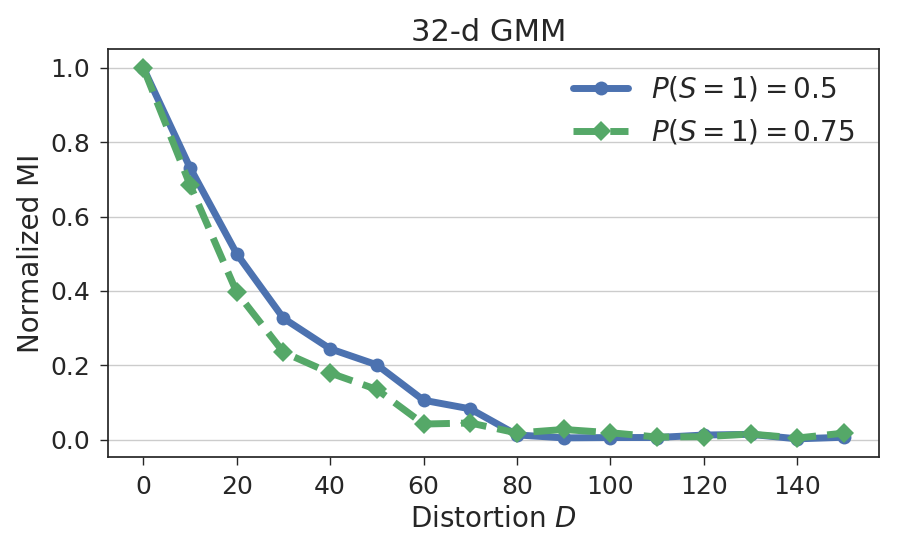}
		\caption{Estimated mutual information between $S$ and $\hat{X}$}
		\label{fig:est-MI-Xr-S-GMM}
	\end{subfigure}
	\caption{Performance of the FUR framework for GMMs.}%
	\label{fig:gaussianperformance}
\end{figure}

Fig. \ref{fig:gaussianperformance} illustrates the performance of the learned FUR scheme against a strong theoretical MAP adversary for a 32-dimensional Gaussian mixture model for both $P(S = 1)=0.75$ and $0.5$. We observe that the {inference} accuracy of the MAP adversary decreases as the distortion increases and asymptotically approaches (as expected) the prior $P(S = 1)$. The encoder obtained via the data-driven approach performs very well when pitted against the MAP adversary (maximum accuracy difference around $0.7 \%$ compared to the theoretical optimal). As another measure of censoring, we estimate the MI of $X_r$ and S using $k$-nearest neighbor method as detailed below. Normalizing it with its theoretical maximum $I(X;S)$ (i.e., with $X_r=X$ for $D=0$), in Fig.~\ref{fig:est-MI-Xr-S-GMM}, we show that MI $\hat{I}(X_r;S)/I(X;S)$ decreases, as expected, when the distortion increases. In other words, for Gaussian mixture data with binary $S$, the data-driven FUR formulation can learn decorrelation schemes that perform as well as those computed under the game-theoretical FUR formulation where the generative decorrelator has access to the data statistics. 

{
We estimate MI using the $k$-nearest neighbor method~\cite{kraskov2004estimating}; in particular, for $n$ $d$-dimensional FUR outputs $X_r$, we first 
estimate the entropy $\hat{H}(X_r)$ as
\begin{align}\label{eq:k-nn-entropy-est}
\hat{H}(X_r) = \psi(N) - \psi(k) + \log(c_d) + \frac{d}{n} \sum_{i=1}^n \log r_i 
\end{align}
where $r_i$ is the distance of the $i$-th sample $\hat{x}_i$ to its $k$-th nearest neighbor, $\psi$ is the digamma function (logarithmic derivative of the gamma function $\Gamma(\cdot)$), and $c_d = \frac{\pi^{d/2}}{\Gamma(1+d/2)}$. 
We then calculate MI as $\hat{I}(X_r; S) = \hat{H}(X_r) - P(S=1)\hat{H}(X_r|S=1) 
- P(S=0)\hat{H}(X_r|S=0)$, 
where $P(S=1)$ and $P(S=0)$ are empirically estimated.}


\vspace{-0.1in}
\section{FUR  for Publicly Available Datasets}
\label{sec:genkisim}
We apply our FUR  framework to four real-world datasets, namely,
UCI Adult~\cite{kohavi1996scaling},
UTKFace~\cite{zhifei2017cvpr},
GENKI~\cite{whitehill2012discriminately}, and HAR~\cite{anguita2013public}, briefly described below. {For all four datasets, we restrict the architecture of $h$, $g$, and the downstream predictive models to neural networks. Note that for tabular datasets (e.g., UCI, HAR), boosting methods including decision trees or support vector models achieve at least comparable predictive performance \cite{kohavi1996scaling,anguita2012human} but are out of scope of this work}. 

\noindent (i) The UCI Adult dataset \cite{kohavi1996scaling}
consists of $10$ categorical and $4$ continuous features and is used to predict a binary salary label (1: salary $> 50$k or  0: salary $\le 50$k). We choose gender or the tuple (gender, relationship) as the sensitive $S$, the remaining features except salary as non-sensitive $X$ (Table SIV in the supplement lists all features), and salary as the target $Y$.

\noindent (ii) The UTKFace dataset \cite{zhifei2017cvpr} 
consists of more than $20$k $200\times 200$ color images of faces labeled by age, ethnicity, and gender. Individuals in the dataset have ages from $0$ to $116$ years and are divided into $5$ ethnicities: White, Black, Asian, Indian, and others including Hispanic, Latino and Middle Eastern.
We take gender as $S$, the image as $X$, and age and ethnicity as two target labels $Y$. Further, we also restrict the data to contain images for ages between 10 and 65.

\noindent{(iii) The GENKI dataset~\cite{whitehill2012discriminately} consists of $1,740$ training and $200$ test samples. Each data sample is a $16\times16$ greyscale face image with varying facial expressions.
We choose gender 
as $S$ and the image as $X$.}

\noindent{(iv) The HAR dataset~\cite{anguita2013public} consists of $561$ features of motion sensor data collected by a smartphone from 30 subjects performing six activities (walking, walking upstairs, walking downstairs, sitting, standing, laying). Each feature is normalized between $-1$ and $1$. We choose subject identity as $S$ and the features of motion sensor data as $X$.}

We use the accuracy of predicting $S$ as the measure of censoring. We evaluate the fairness guarantees of $X_r$ by computing the DemP obtained on tasks using $Y$. To this end, 
we compute the following maximal difference as a proxy for DemP in Definition \ref{Def:ThreeFairnessMeasures} (includes non-binary $Y$ and $S$):
\thickmuskip=2mu
\begin{align}\label{eq:statistic_DP}
    \Delta_{\text{DemP}}(y)=\max_{s,s'\in \mathcal{S}} |P(\hat{Y}=y|S=s)-P(\hat{Y}=y|S=s')|
\end{align}
with smaller values of $\Delta_{\text{DemP}}(y)$ suggesting better DemP fairness guarantees. For binary $Y$, $\Delta_{\text{DemP}}(y)$ in \eqref{eq:statistic_DP} simplifies to a single value that we denote as $\Delta_{\text{DemP}}$. {In our experiments, we use the empirical frequencies to estimate $P(\hat{Y}=y|S=s)$ for a chosen $(y,s)$.} We illustrate both censoring and fairness results for the abovementioned datasets in the following subsections. Experimental and model details are in the supplement. 

\vspace{-0.15in}
\subsection{Illustration of Results for UCI Adult Dataset}
\label{sec:uci_adult}
For the UCI Adult dataset with both categorical and continuous features as shown in Table SIV in the supplement, we consider two cases: \\(i) Case I: binary $S$ by choosing `gender' as sensitive feature  \\ 
(ii) Case II: non-binary $S$ by considering both `gender' and `relationship' as sensitive. For UCI, `relationship' has $6$ distinct values, and therefore, $S$ has $12$ possibilities. \\
For both cases, `salary' is the binary target $Y\in\{0,1\}$, with $Y=1$ denoting salary $>50K$. 
Since the two values for $\Delta_{\text{DemP}}(y)$ in \eqref{eq:statistic_DP} are the same for binary $Y$, we write $\Delta_{\text{DemP}}$ when illustrating results. 

\noindent\textbf{Case I: Binary Sensitive Feature.}

\begin{figure}[h!]
	\centering
	\begin{subfigure}[t]{0.45\textwidth}
		\includegraphics[width=1\columnwidth,trim=15 10 5 37,clip]{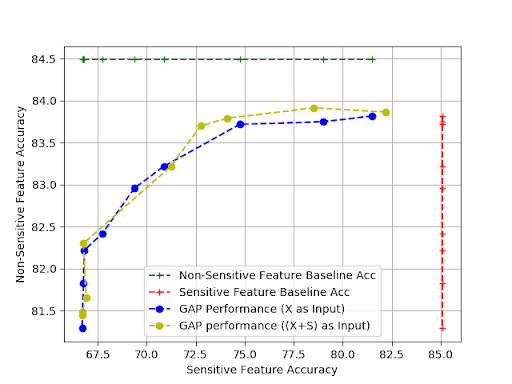}
		\caption{Salary vs. gender classification accuracy }
		\label{fig:UCI-CaseI-Censor}
	\end{subfigure}
	\begin{subfigure}[t]{0.46\textwidth}
		\includegraphics[width=1\columnwidth,trim=10 2 5 15,clip]{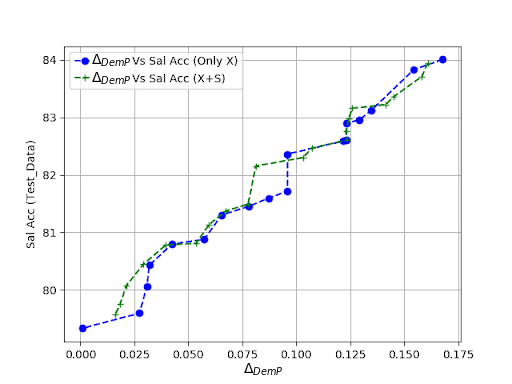}
		\caption{Salary classification accuracy vs. $\Delta_{\rm{DemP}}$}
		\label{fig:UCI-CaseI-fair}
	\end{subfigure}
	\caption{Results for UCI Adult: Case I. In Fig. \ref{fig:UCI-CaseI-Censor}, 
	the green and red lines denote the baseline performances for the target $Y$ (salary) and sensitive $S$ (gender), respectively; in Fig. \ref{fig:UCI-CaseI-fair}, the value of $\Delta_{\text{DemP}}$ for the original test data is $0.2$. {In both plots, each point corresponds to a specific value of achieved test distortion; for Figs. \ref{fig:UCI-CaseI-Censor} and \ref{fig:UCI-CaseI-fair}, the achieved test distortion for the blue points ranges over $(0.69,4.1)$  and $(0.69,4.4)$, respectively, with decreasing distortion from left to right for each plot. The achieved test distortion for the yellow-green points ranges over $(0.87,4.2)$ and $(0.87,4.9)$, respectively.} }%
	\label{fig:UCI-CaseI}
\end{figure}

Fig. \ref{fig:UCI-CaseI} illustrates the censoring and fairness performance of the generated $X_r$ for the UCI dataset. For censoring, the performance is evaluated via the tradeoff between the classification accuracies of salary (utility of $X_r$) and gender (censoring of $S$). {Note that salary accuracy is evaluated as a downstream task via a separately learned classifier that uses $X_r$ while gender accuracy is a measure of performance of the neural network adversary $h$ in the FUR model}. We evaluate fairness via the tradeoff between salary accuracy and $\Delta_{\text{DemP}}$. We consider two possible inputs to the encoder $g(\cdot)$ in \eqref{eq:generalopt}, i.e., only $X$ or both $(X,S)$. 

%

As illustrated in Fig. \ref{fig:UCI-CaseI-Censor}, the baseline\footnote{The baseline performances are the salary and gender accuracies as well as $\Delta_{\text{DemP}}$ obtained from the original uncensored test dataset.} salary and gender accuracies for the UCI dataset are about $84.5\%$ and $85\%$, respectively. Further, for the FUR $X_r$ and downstream $\hat{Y}$: \\
(i) the smallest gender accuracy achievable is about $66\%$, $20\%$ below its baseline, while the lowest salary accuracy is about $82\%$, $2.5\%$ below its baseline. Since the likelihood of a male in the original test data is $66\%$, with increasing distortion, the FUR gender accuracy is as good as a random guess, i.e., the generated $X_r$ hides gender effectively while maintaining high salary accuracy. \\
(ii) For the same gender accuracy, using both $(X,S)$ seems useful only in the high utility setting (salary accuracy $\ge 83\%$). 
From Fig. \ref{fig:UCI-CaseI-fair}, we make the following two observations:

 \noindent (i) salary classification accuracy and $\Delta_{\text{DemP}}$ have an approximately affine relationship, and when $\Delta_{\text{DemP}}\approx 0$, the salary accuracy is $\ge 79\%$, i.e., the FUR framework is effective in approaching perfect DemP with a small reduction in utility; \\(ii) the FURs $X_r$ generated from either $X$ or $(S,X)$ lead to similar fairness guarantees.
For $\Delta_{\text{DemP}}=0.06$, state-of-the-art approaches in \cite{edwards_Storkey2016} and \cite{Madras_2018} achieve $2\%$ and $2.5\%$ higher salary accuracy than ours, respectively;
however, 
our approach is distinct in achieving 
$\Delta_{\text{DemP}}\approx 0$ with salary accuracy $\ge 79\%$. 

{From Fig. \ref{fig:UCI-CaseI-Censor}, we see that gender accuracy saturates at 67\% while achieving a salary accuracy of at least 81\% for a specific value of distortion bound $D$, and therefore, test distortion; in turn, this choice of $D$ corresponds in Fig. \ref{fig:UCI-CaseI-fair} to $\Delta_{\text{DemP}}\approx 0.06$. Further reducing $\Delta_{\text{DemP}}$ requires further increasing $D$, thus lowering the salary accuracy to 79\% for $\Delta_{\text{DemP}}\approx 0$. This is because classification accuracy captures an average measure of correctness and is dominated by the performance over the majority class. On the other hand, $\Delta_{\text{DemP}}$ captures the difference in performance of the intended classifier on each of the two classes. Thus, enforcing fairness via $\Delta_{\text{DemP}}$ reduces salary accuracy thereby highlighting the tradeoff between guaranteeing fairness and utility.}



We can also evaluate the fairness performance of the generated $X_r$ by using the EO measure in Definition \ref{Def:ThreeFairnessMeasures}. Thus, for $Y \in \{0,1\}$ where $Y=1$ when salary $>50K$, $S \in \{0,1\}$ (female:1 and male:0), and $\hat{Y}\in \{0,1\}$, we write $\Delta_{\text{EO}}(y), \forall y\in\{0,1\}$ as:
\thickmuskip=1mu
\begin{equation}
    \Delta_{\text{EO}}(y)\triangleq\left|P(\hat{Y}=y|S=0,Y=y) -P(\hat{Y}=y|S=1,Y=y)\right|.
    \label{eq:measure_EO1}
\end{equation}
\thickmuskip=3mu

\begin{figure}
    \centering
    \includegraphics[width=3in]{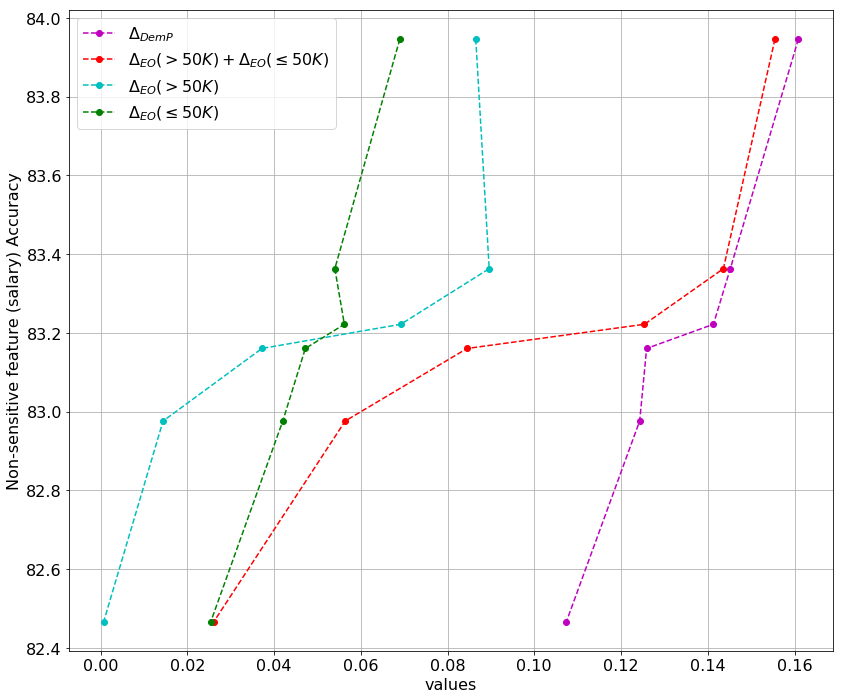}
    \caption{Evaluation of equalized odds fairness metric under Case I for the UCI Adult dataset. The EO measures $\Delta_{\text{EO}}(1)$ and $\Delta_{\text{EO}}(0)$ are defined in \eqref{eq:measure_EO1}. The red curve plotting $\Delta_{\text{EO}}=\Delta_{\text{EO}}(1)+\Delta_{\text{EO}}(0)$ matches $\Delta_{\text{EO}}$ in Figure 2(b) of \cite{Madras_2018}. {Each point corresponds to a specific value of achieved test distortion  ranging over $(0.59,2.01)$, with distortion decreasing from the left to the right for each plot.}
    }
    \label{fig:UCI-CaseI-EO}
\end{figure}

{Note that for binary $Y$, as is the case here, \eqref{eq:measure_EO1} is the same as the definition of EO in \eqref{eq:equalied_odds}.} 
From Fig. \ref{fig:UCI-CaseI-EO}, which plots salary accuracy vs. DemP or EO measures of fairness, we observe that while the salary accuracy is above $82.4\%$, 
the values of $\Delta_{\text{EO}}(1)$ and $\Delta_{\text{EO}}(0)$ decrease to $0.0007$ and $0.0254$, respectively. 
To understand the significance of these results, we compare against the state-of-the-art in~\cite{Madras_2018}, wherein fair salary classifiers for both DemP and EO measures, referred to as LAFTR-DP\footnote{Learned Adversarially Fair and Transferable Representations (LAFTR)} and LAFTR-EO, respectively, are learned for the UCI dataset. For the LAFTR-DP, the authors also compute the resulting EO of the DemP classifier. {As a preamble to the following comparisons, we note that fair predictors, trained on specific tasks, will do at least as well as the same predictors learned on fair representations.}

We make the following observations: (i) when $\Delta_{\text{EO}}(1)+\Delta_{\text{EO}}(0)=0.04$\footnote{\cite{Madras_2018} introduced an EO measure as $\Delta_{\text{EO}}\triangleq\Delta_{\text{EO}}(1)+\Delta_{\text{EO}}(0)$},
our salary accuracy is $1.3\%$ smaller than that achieved by LAFTR-DP (cf. Fig. 2(b) in \cite{Madras_2018}), but our minimal achieved value of $\Delta_{\text{EO}}(1)+\Delta_{\text{EO}}(0)$ is only $72\%$ of that achieved by LAFTR-DP and is the same as the value achieved by LAFTR-EO, which uses EO as the fairness metric to train a salary classifier; (ii) the decrease of $\Delta_{\text{EO}}(1)+\Delta_{\text{EO}}(0)$ is even larger than $\Delta_{\text{DemP}}$. That is, even though the representation is generated to satisfy DemP, it can also provide competitive downstream EO fairness guarantees. This, in turn, justifies the rationality of generating fair representations under DemP. 

\noindent\textbf{Case II: Non-binary Sensitive Feature.}

\begin{figure*}[h]
	\centering
	\begin{subfigure}[b]{0.325\textwidth}
		\includegraphics[height=4.5cm, width=1\columnwidth,trim=20 5 5 0,clip]{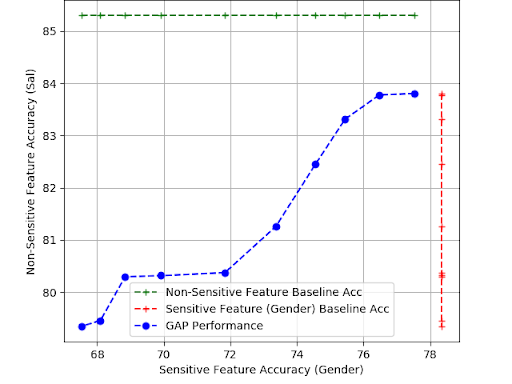}
		\caption{Gender}
		\label{fig:UCI-CaseII-Censor1}
	\end{subfigure}
	\begin{subfigure}[b]{0.325\textwidth}
		\includegraphics[height=4.5cm, width=1\columnwidth,trim=20 10 5 0,clip]{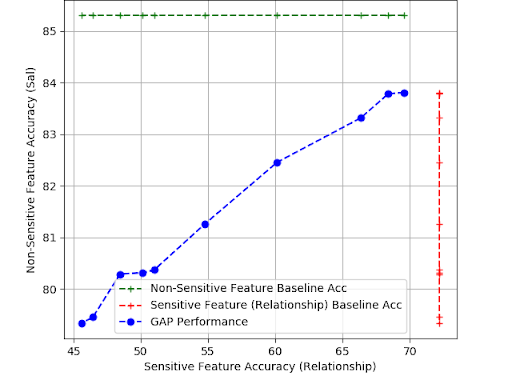}
		\caption{Relationship}
		\label{fig:UCI-CaseII-Censor2}
	\end{subfigure}
	\begin{subfigure}[b]{0.325\textwidth}
		\includegraphics[height=4.5cm, width=1\columnwidth,trim=3 10 0 40,clip]{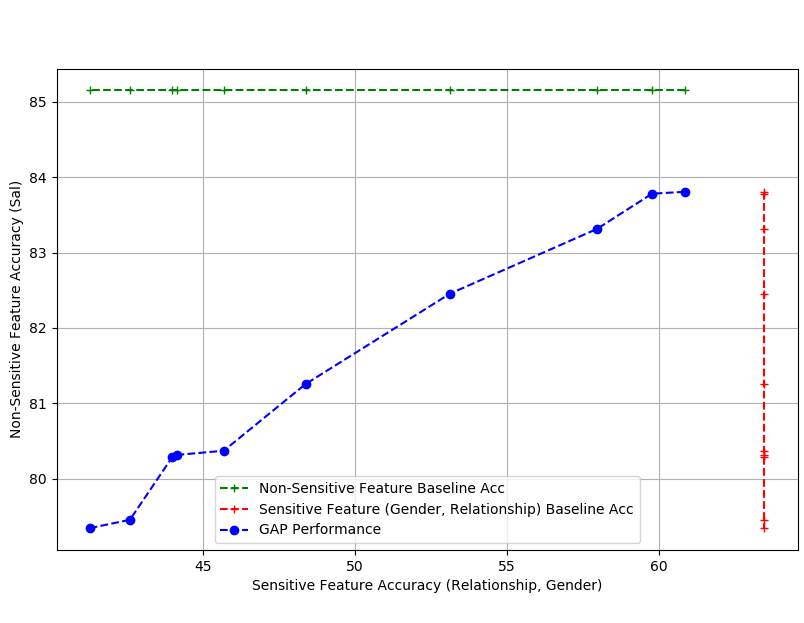}
		\caption{ (Gender, Relationship)}
		\label{fig:UCI-CaseII-Censor3}
	\end{subfigure}
	\caption{Tradeoff between classification accuracy of non-sensitive feature (salary) and sensitive features (gender and/or relationship) under Case II for the UCI Adult dataset. Note that we use the classification accuracy obtained from the original testing dataset as the baseline performance, which is denoted by the green and red lines for the target variable (salary) and the sensitive variable (gender or/and relationship), respectively. {In every plot, each point corresponds to a specific value of achieved test distortion (over all features except gender and relationship) ranging over $(0.58,2.1)$, with distortion decreasing from the left to the right for each plot.} }%
	\label{fig:UCI-CaseII-Censoring}
\end{figure*}

\begin{figure}[h!]
	\centering
	\begin{subfigure}[t]{0.4\textwidth}
		\includegraphics[width=1\columnwidth,trim=10 7 5 37,clip]{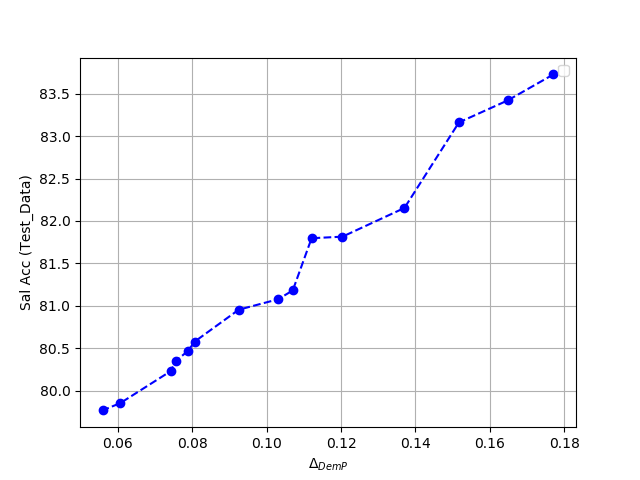}
		\caption{Gender}
		\label{fig:UCI-CaseII-DemP1}
	\end{subfigure}
	\begin{subfigure}[t]{0.4\textwidth}
		\includegraphics[width=1\columnwidth,trim=10 7 5 37,clip]{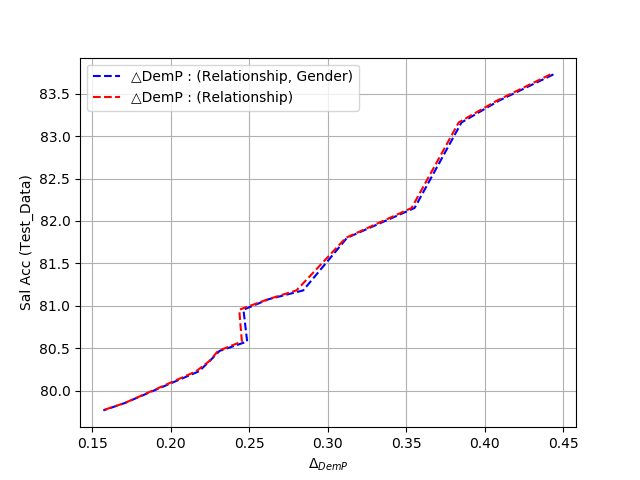}
		\caption{Relationship (or) (Relationship, Gender)}
		\label{fig:UCI-CaseII-DemP2}
	\end{subfigure}
	\caption{Case II for UCI Adult: Tradeoffs between salary accuracy and the $\Delta_{\rm{DemP}}$ of gender and/or relationship.
    For the original test data, $\Delta_{\text{DemP}}$ for gender, relationship and the pair (gender, relationship) is $0.2$, $ 0.438$ and $0.443$, respectively. {In every plot, each point corresponds to a specific value of achieved test distortion (over all features except gender and relationship) ranging over $(0.58,2.1)$, with distortion decreasing from the left to the right for each plot. 
    }}
    \vspace{-0.2in}
\label{fig:UCI-CaseII-fairness}
\end{figure}

Figs. \ref{fig:UCI-CaseII-Censoring} and \ref{fig:UCI-CaseII-fairness} illustrate the censoring and fairness performances of the generated $X_r$ in hiding `gender' and `relationship', respectively, while preserving `salary' information. 
Fig. \ref{fig:UCI-CaseII-Censoring} illustrates the tradeoff between salary and sensitive feature $S$ accuracies when $S$ is either gender, or relationship, or both. 
From Fig. \ref{fig:UCI-CaseII-Censoring}, we observe that while the salary accuracy is above $79\%$, the classification accuracies of gender and/or relationship are about $66\%$ (Fig. \ref{fig:UCI-CaseII-Censor1}), $45\%$ (Fig. \ref{fig:UCI-CaseII-Censor2}) and $41\%$ (Fig. \ref{fig:UCI-CaseII-Censor3}), respectively. Note that the probabilities of male, husband, and the combination (male, husband) are $66\%$, $40\%$ and $40\%$, respectively, in the original test data. Therefore, while the salary accuracy is preserved at $79\%$, the inferences of gender, relationship, and combination (gender, relationship) approach random guessing with these priors. Thus, our FUR framework can effectively hide one or more sensitive features. However, suppressing multiple correlated sensitive features comes at a cost of a reduction in salary accuracy. 
Thus, comparing Figs. \ref{fig:UCI-CaseI-Censor} and \ref{fig:UCI-CaseII-Censor1}, we see a maximal reduction of $3\%$ in salary accuracy for a given gender accuracy\footnote{\label{ref:footnote9}In Figs., \ref{fig:UCI-CaseI-Censor} and \ref{fig:UCI-CaseII-Censor1}, the baseline performances are different because for Case II, the feature variable $X$ does not contain `relationship'.}.

For Case II, Figs. \ref{fig:UCI-CaseII-DemP1} and \ref{fig:UCI-CaseII-DemP2} illustrate the tradeoffs between the salary accuracy and $\Delta_{\text{DemP}}$ for $S$ chosen as gender or relationship or both.
We observe that while salary accuracy is above $94\%$ of the baseline performance, 
the value of $\Delta_{\text{DemP}}$ is dropped to $25\%$ for gender and to about $34\%$ for both relationship and their combination. In short, $X_r$ works well in decorrelating gender and relationship both separately and jointly without affecting downstream classifier performance. From Fig. \ref{fig:UCI-CaseII-DemP2}, we observe that the value of $\Delta_{\text{DemP}}$ for the combination is almost the same as that for relationship. {In addition, comparing the results in Figs. \ref{fig:UCI-CaseI-fair} and \ref{fig:UCI-CaseII-DemP1}, for any given $\Delta_{\text{DemP}}$ for gender, the salary accuracy in Case II is about $1\%$ lower
than that in Case I; this can be viewed as the cost of {eliminating} {a potentially sensitive feature (relationship) that is also correlated with the target feature (see also, footnote \ref{ref:footnote9})}.
Finally, comparing the results in Figs. \ref{fig:UCI-CaseI-fair} and \ref{fig:UCI-CaseII-DemP2}, for any given salary accuracy, $\Delta_{\text{DemP}}$ for gender in Case II is about {$0.25$ }
 higher
than that in Case I; this can be viewed as the effect of using non-binary sensitive features on $\Delta_{\text{DemP}}$, now defined as the maximum over all values taken by the non-binary sensitive feature.}


\vspace{-0.15in}
\subsection{Illustration of Results for UTKFace Dataset}
In the UTKFace dataset, the face images are the non-sensitive $X$. We choose `gender' as the sensitive $S$; focusing on multiple downstream tasks, we consider both ethnicity classification and age regression, for which we choose `ethnicity' or `age' as the target variable $Y$, respectively. 
For the two downstream applications, the corresponding supports of $Y$ are $\mathcal{Y}=\{\rm{White},\, \rm{Black},\, \rm{Asian},\, \rm{Indian}\}$ and  $\mathcal{Y}=\{i\in \mathbb{Z}: 10\leq i\leq 65\}=[10,65]$, respectively. We use the maximum of the DemP measure (defined in \eqref{eq:statistic_DP}) over the support $ Y$, i.e., $\Delta_{\rm{DemP}}= \max_{y\in \mathcal{Y}}\Delta_{\rm{DemP}}(y)$, as the achieved fairness level.

\begin{figure*}[h]
    \centering
    \includegraphics[width=6.5in,trim=8 6 5 8,clip]{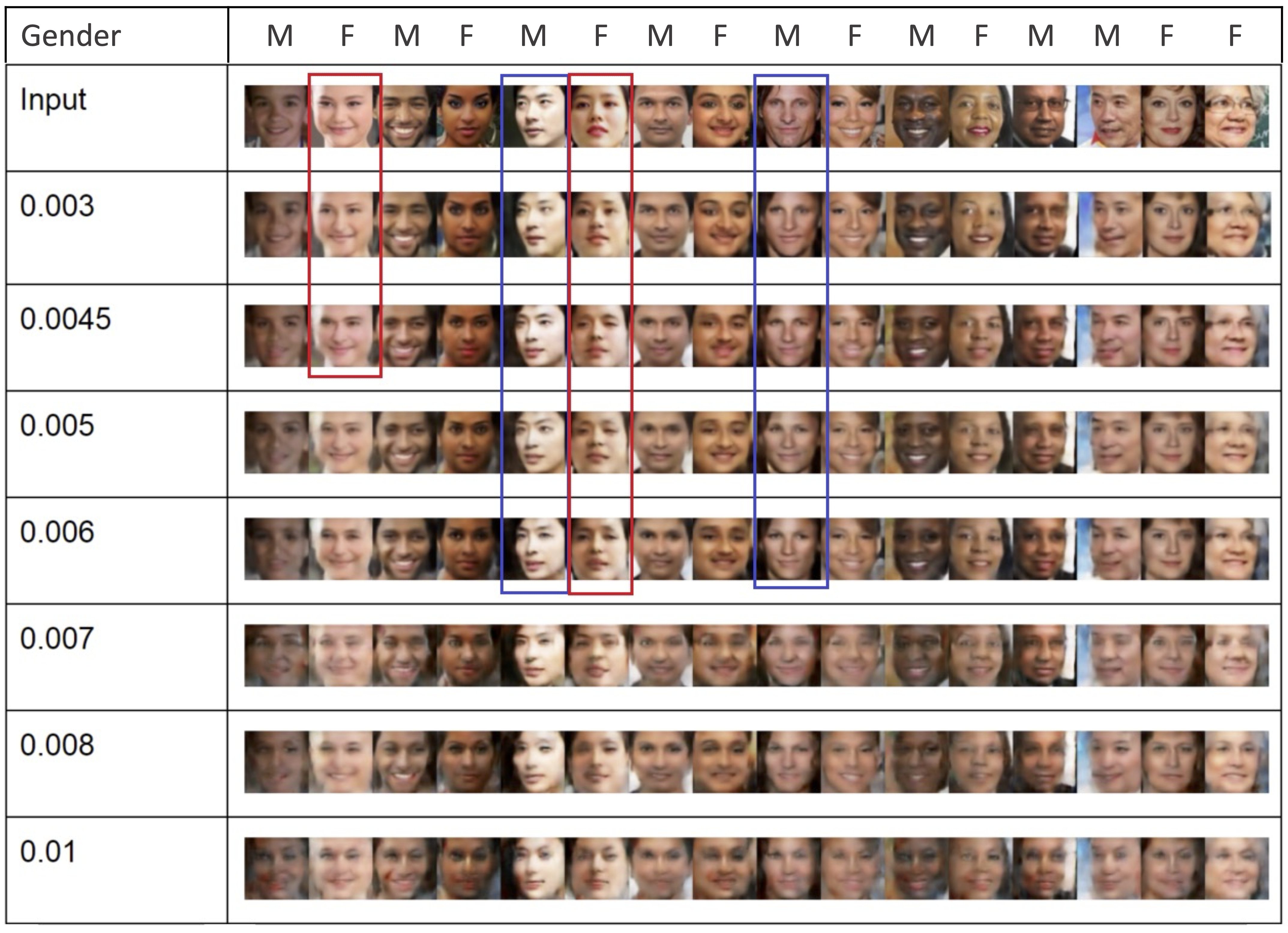}
    \caption{The encoded face images for different values of per-pixel distortions for the UTKFace dataset. Set of vertical faces highlighted in boxes makes explicit how the sensitive feature (gender) is changed with increasing distortion. {The ground truth gender values for the images are shown in the top-most row}.}
    \label{fig:GAP_UTK}
\end{figure*}

Fig.~\ref{fig:GAP_UTK} illustrates the output $X_r$ for $16$ typical\footnote{The $16$ typical faces covers the $8$ possible combination of $2$ gender (male and female) and $4$ ethnicities (White, Black, Asian and Indian) and includes young, adult and old faces.} faces in the UTKFace dataset for increasing per-pixel distortion.  From Fig.~\ref{fig:GAP_UTK}, we observe that: (i) for a small per-pixel distortion (e.g., $0.003$), gender-distinguishing features such as lip color are smoothed out; and (ii) at higher per-pixel distortion (e.g., $0.006$), the FUR framework can generate a face with an opposite gender (see the highlighted examples in Fig.~\ref{fig:GAP_UTK}) thereby completely obfuscating this sensitive feature; 
(iii) when the average per-pixel distortion is too large (e.g., $0.01$), the representations generated are often too blurred. 

\begin{figure}[h]
	\centering
	\begin{subfigure}[t]{0.4\textwidth}
		\includegraphics[width=1\columnwidth,trim=0 6 5 10,clip]{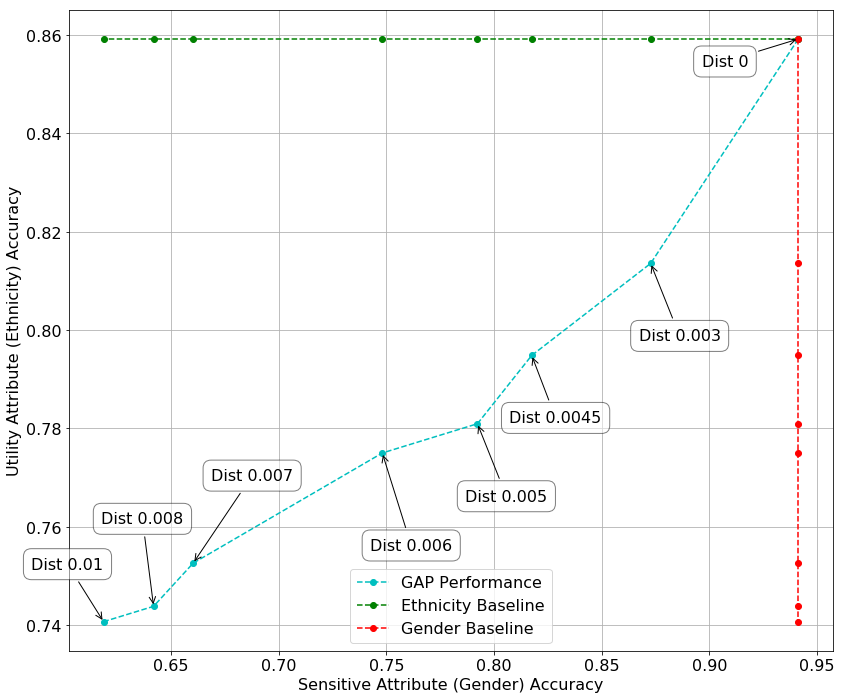}
		\caption{Ethnicity vs. gender classification accuracy}
		\label{fig:UTK_race_censor}
	\end{subfigure}
	\begin{subfigure}[t]{0.4\textwidth}
		\includegraphics[width=1\columnwidth,trim=0 6 5 10,clip]{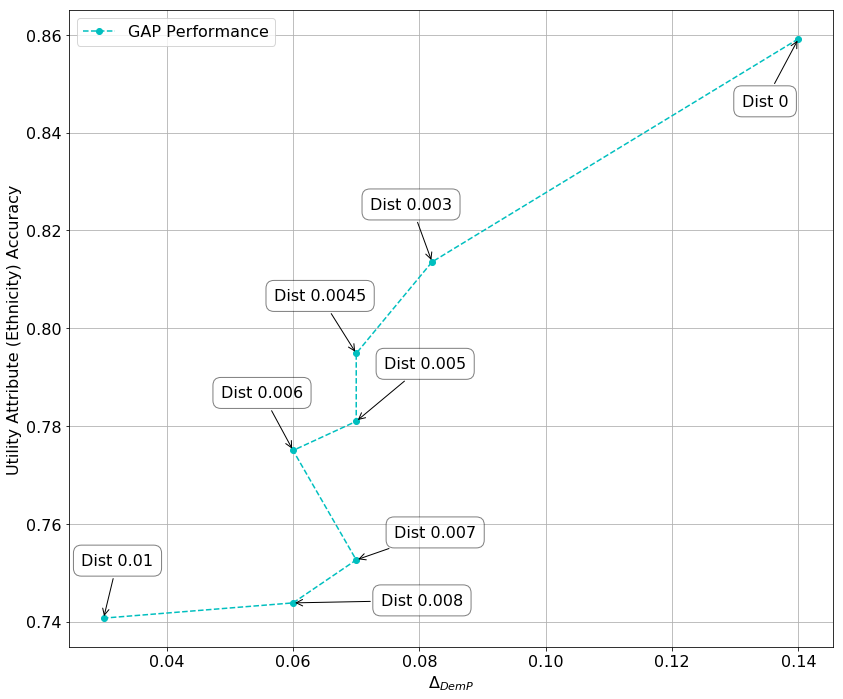}
		\caption{Ethnicity classification accuracy vs. $\Delta_{\rm{DemP}}$}
		\label{fig:UTK_race_fairness}
	\end{subfigure}
	\caption{Ethnicity classification accuracy vs. gender classification and $\Delta_{\rm{DemP}}$ for the UTKFace dataset. In Fig. \ref{fig:UTK_race_fairness}, the x-axis is the maximal value of DemP in \eqref{eq:statistic_DP}  over the four ethnicities and `dist' indicates the per pixel distortion.    }%
	\vspace{-0.1in}
	\label{fig:UTK_race}
\end{figure}

\begin{figure}[h]
	\centering
	\begin{subfigure}{0.4\textwidth}
		\includegraphics[width=1\columnwidth,trim=5 6 5 6,clip]{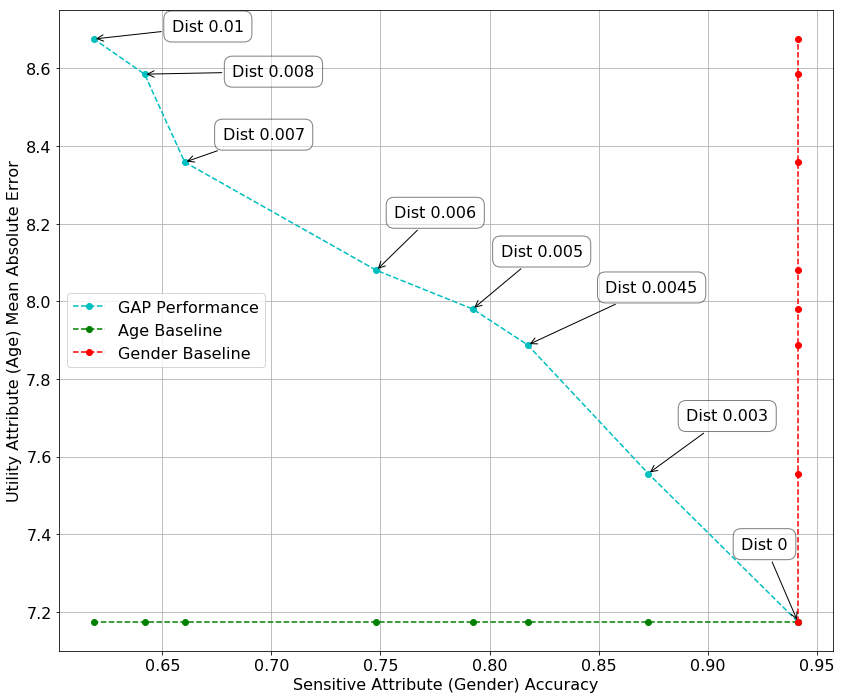}
		\caption{Mean absolute error of age prediction vs. gender classification accuracy  }
		\label{fig:UTK_age_censor}
	\end{subfigure}
	\begin{subfigure}{0.4\textwidth}
		\includegraphics[width=1\columnwidth,trim=5 6 5 6,clip]{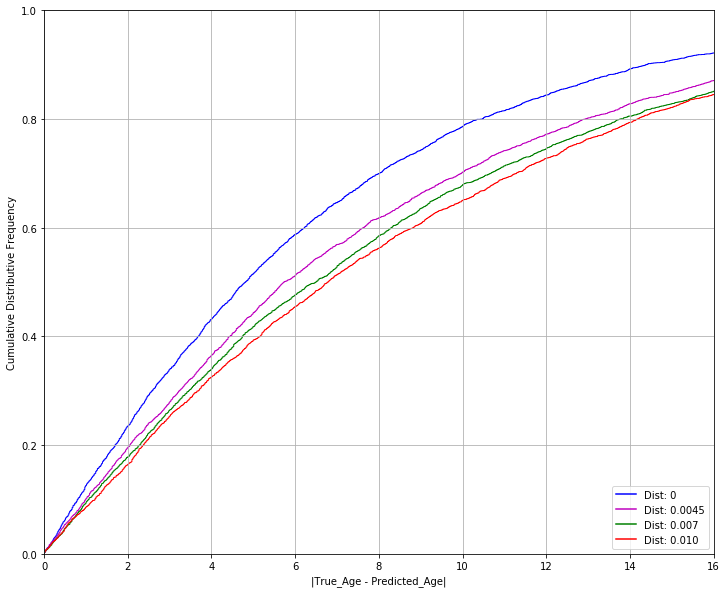}
		\caption{The CDF of the difference between the true and predicted age}
		\label{fig:UTK_age_cdf}
	\end{subfigure}
	\caption{Utility of age regression on the UTKFace dataset. Note that `dist' indicates the per pixel distortion. }%
	\label{fig:UTK_age1}
	\vspace{-0.18in}
\end{figure}

Figs.~\ref{fig:UTK_race_censor} and \ref{fig:UTK_age1} show the tradeoffs between gender classification accuracy and appropriate measures for ethnicity classification and age regression, respectively. In Fig. \ref{fig:UTK_race_censor}, while gender classification accuracy is about $62\%$ and decreases about $35\%$ from the baseline performance, the classification accuracy of ethnicity is above $74\%$ and only decreases $14\%$ from its baseline performance. Note that in the original testing data, the highest marginal probabilities for gender and ethnicity are $54.6\%$ (likelihood of male) and $43.2\%$ (likelihood of White), respectively. Therefore, gender accuracy is better than a random guess by only $7.4\%$ while ethnicity accuracy is better than a random guess by $30.8\%$, i.e., the generated $X_r$ hides gender information well while maintaining ethnicity. For age regression, we use the mean absolute error (MAE), i.e., the average absolute difference between the predicted age and the true age, as the utility measure. In Fig. \ref{fig:UTK_age_censor}, we observe that while the classification accuracy for gender is about $62\%$, which is a $35\%$ decrease from the baseline performance of $94\%$, the increase in the MAE is $1.5$ which is about a $20\%$ increase from the baseline performance of $7.2$ years. Fig. \ref{fig:UTK_age_cdf} shows the cumulative distribution function (CDF) of the difference between the true and predicted age for various distortions, from which we can see that the drop of the cumulative probability is at most $1\%$. Thus, the generated FUR guarantees reliable performance for both age and ethnicity prediction; thus, constraining the distortion of the generated $X_r$ can be effective in guaranteeing utility for multiple tasks.

\begin{figure}[h]
	\centering
	\begin{subfigure}[t]{0.4\textwidth}
		\includegraphics[width=1\columnwidth]{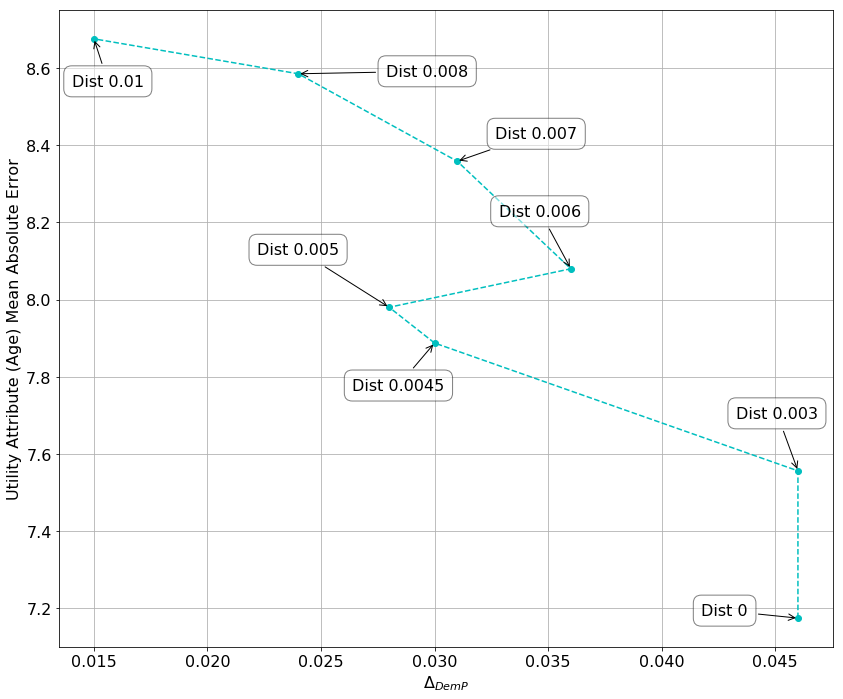}
		\caption{Mean absolute error of age prediction vs. $\Delta_{\rm{DemP}}$ }
		\label{fig:UTK_age_fair}
	\end{subfigure}
	\begin{subfigure}[t]{0.4\textwidth}
		\includegraphics[width=1\columnwidth]{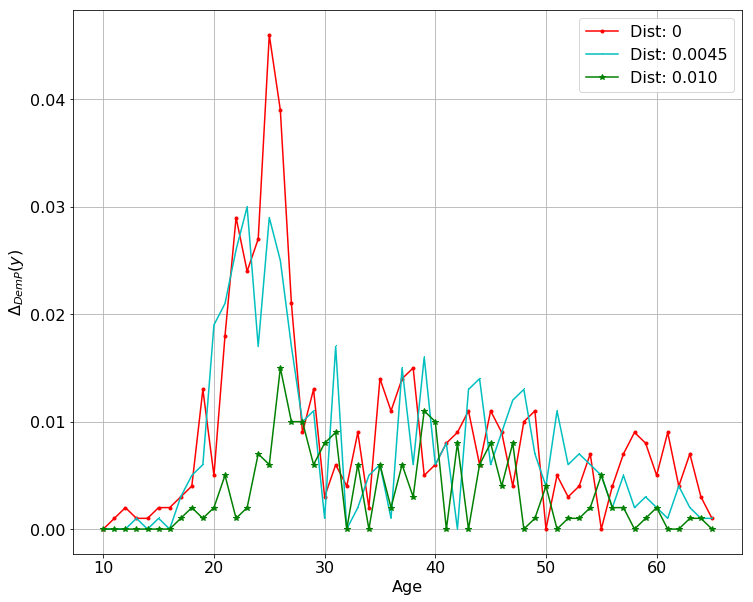}
		\caption{The demographic parity measure for various distortions}
		\label{fig:UTK_age_DemPdetail}
	\end{subfigure}
	\caption{Achieved demographic parity for the age regression task on the UTKFace dataset. Note that in Fig. \ref{fig:UTK_age_fair}, the x-axis is the maximal value of DemP in \eqref{eq:statistic_DP}  over the chosen age range (10-65) and `dist' indicates the per pixel distortion.  }%
	\label{fig:UTK_age2}
\end{figure}

In Figs.~\ref{fig:UTK_race_fairness} and \ref{fig:UTK_age2}, we illustrate the tradeoff between the utility measure and $\Delta_{\rm{DemP}}$ of the generated $X_r$ in ethnicity classification and age regression, respectively. In Fig.~\ref{fig:UTK_race_fairness}, we observe that while achieving about $86\%$ of the baseline classification accuracy, the $\Delta_{\rm{DemP}}$ is reduced
to $0.03$, which is $20\%$ of the $\Delta_{\rm{DemP}}=0.14$ in the original testing data. 
Table \ref{tab:my_label} shows the decrease of $\Delta_{\rm{DemP}}$  for each of the four ethnicities as the distortion increases.  
In Fig.~\ref{fig:UTK_age_fair}, while preserving $86\%$ of the utility baseline performance, the $\Delta_{\rm{DemP}}$, i.e., the maximal value of demographic parity measure over the $56$ age values, decreases to $0.015$, which is less than $33\%$ of the $\Delta_{\rm{DemP}}=0.046$ in the original testing data. Fig.~\ref{fig:UTK_age_DemPdetail} shows the demographic measure $\Delta_{\rm{DemP}}(y)$ , $y\in[10,65]$, for various distortions; we observe that when the pixel distortion is $0.01$, even while $\Delta_{\rm{DemP}}=0.015$, $\Delta_{\rm{DemP}}(y)=0$ for $17$ distinct ages. That is, the predictions of these $17$ ages are completely independent of gender and DemP is achieved for those predictions.

\definecolor{green}{rgb}{0.9,1,0.9}
\definecolor{LightRed}{rgb}{1, 0.9,0.9}
\definecolor{yellow}{rgb}{1, 1,0.9}
\newcolumntype{g}{>{\columncolor{yellow}}c}
\begin{table*}[h]
    \small
    \centering
    \begin{tabular}{|g|c|c|c|c|c|c|c|c|}
    \hline
    \rowcolor{green}
         Distortion & 0 & 0.003 & 0.0045 & 0.005 & 0.006& 0.007& 0.008& 0.01 \\\hline 
          $\Delta_{\text{DemP}}(\rm{White})$& 0.061& 0.055 & 0.04& 0.03& 0.03& 0.02& 0.02 & 0.01 
          \\ 
          $\Delta_{\text{DemP}}(\rm{Black})$& 0.109& 0.021& 0.02& 0.05& 0.03& 0.05& 0.03& 0.03
          \\ 
          $\Delta_{\text{DemP}}(\rm{Asian})$& 0.14 & 0.082& 0.07& 0.07& 0.06&0.07& 0.06& 0.03
          \\ 
          $\Delta_{\text{DemP}}(\rm{Indian})$& 
           0.031 & 0.006 &0.01 & 0 & 0.01 & 0& 0.01& 0.01 
          \\\hline 
    \end{tabular}
    \caption{Demographic parity fairness (indicated by $\Delta_{\text{DemP}}(\cdot)$) of ethnicity classification on the UTKFace dataset. 
    }
    \label{tab:my_label}
\end{table*}


\vspace{-.1in}
{
\subsection{Illustration of Results for GENKI Dataset}
\label{sec:GENKI}
For the GENKI dataset, we consider the following two approaches to decorrelating the data $(X,S)$: the feedforward neural
network decorrelator (FNND) and the transposed convolution neural network decorrelator (TCNND). Specific architectural details for both can be found in the supplement. Fig.~\ref{fig:genkiprivacy_util} illustrates the gender classification accuracy of the adversary for different values of distortion. It can be seen that the adversary's accuracy of classifying $S$ (gender) decreases
as the distortion increases. Given the same distortion value, the FNND achieves lower gender classification accuracy compared to the TCNND. 
An intuitive explanation for this is that the FNND uses both the noise vector and the original image to generate the processed image, while the TCNND generates the noise mask independently of the original image and then adds this mask to the original image in the final step. }

\vspace{-0.005in}
\noindent{\textbf{Censoring vs. differential privacy guarantees}: For this dataset, in addition to highlighting the role of GAN-like architectures to learn fair representations, we also explore the effect of censoring on assuring privacy of the sensitive (here, gender) feature. 
Differential privacy (DP) has emerged as the gold standard for data privacy \cite{Dwork_DP_Survey}. 
Thus, one way to censor and privatize data is to add noise with differential privacy guarantees \cite{Dwork_DP_Survey}. Since the dataset is continuous valued, we consider two types of additive DP noise mechanisms at the pixel level: Gaussian and Laplacian. We vary the variances of the Laplace and Gaussian noise to then obtain a specific local DP guarantee\footnote{DP, by definition, guarantees that the output of a differentially private (randomizing) mechanism cannot aid in distinguishing between two neighboring (defined appropriately) datasets. Local DP is stronger than DP in that it provides such a guarantee for any pair of inputs.} building on \cite{Kairouz2016}. 
We compute the resulting DP guarantees provided by independent Laplace and Gaussian noise-adding mechanisms for different distortion values. The details are provided in the supplement (see Section SII-B). 
In Fig.~\ref{fig:genkiprivacy_util}, we compare the gender classification accuracy of the learned FUR schemes with those obtained by adding Laplace or Gaussian noise. 
We see that for the same distortion, the learned FUR schemes achieve much lower gender classification accuracy. In Table \ref{tb:genki}, we observe that even when a large amount of noise is added to each pixel, the privacy risk ($\epsilon$) is still significantly high. Furthermore, such noise levels deteriorate the expression classification accuracy (cf. Fig.~\ref{fig:genkiprivacy_util}). It is worth noting that the distortion constraint for the FUR framework is an average over the entire image.
}


\begin{figure}[!hbpt]
	\centering
	\begin{subfigure}[t]{0.46\textwidth}
        \includegraphics[width=1\columnwidth]{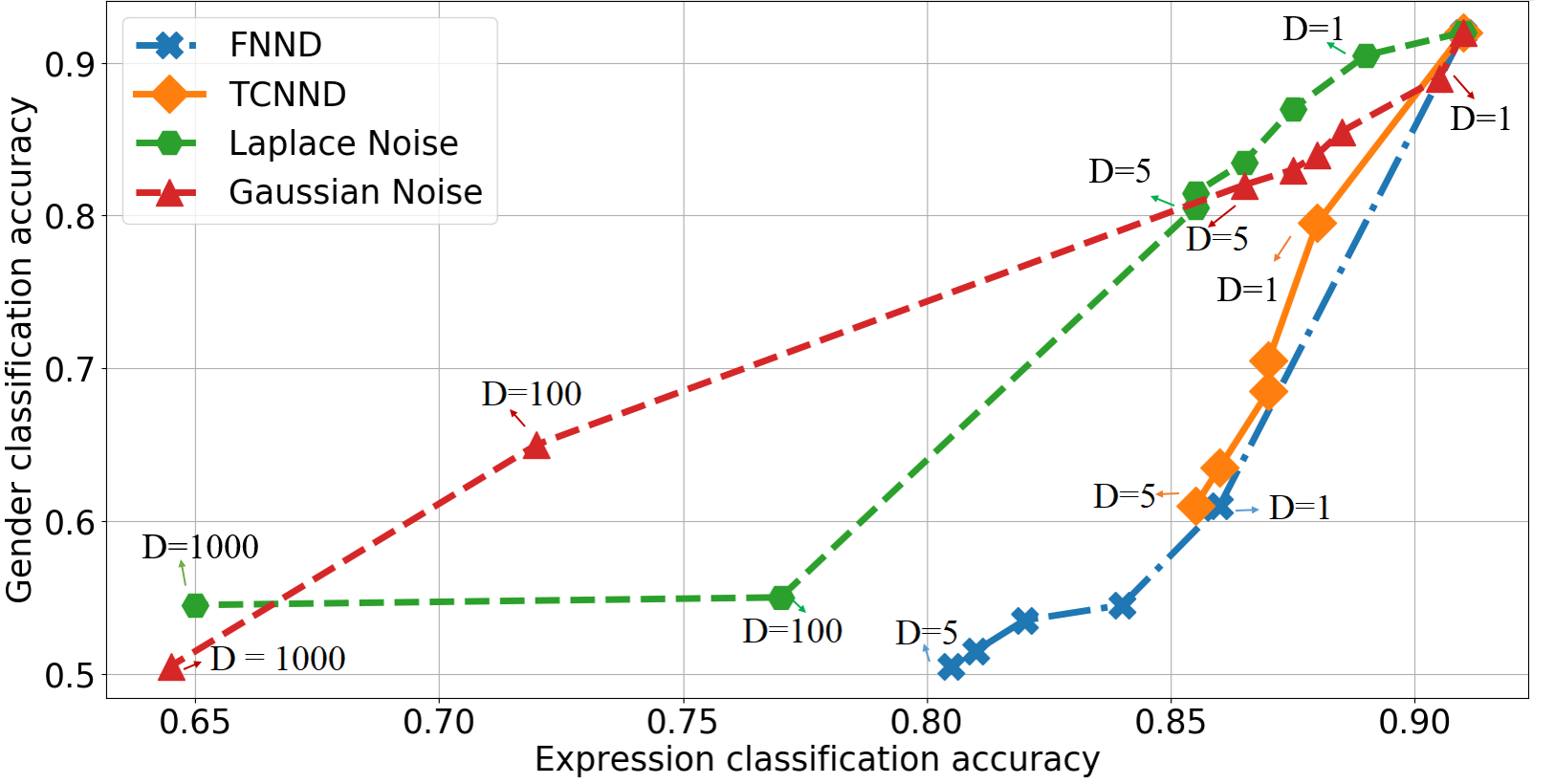}
		\caption{Gender vs. expression classification accuracy}
		\label{fig:genkiprivacy_util}
	\end{subfigure}
	\begin{subfigure}[t]{0.45\textwidth}
        \includegraphics[width=1\columnwidth]{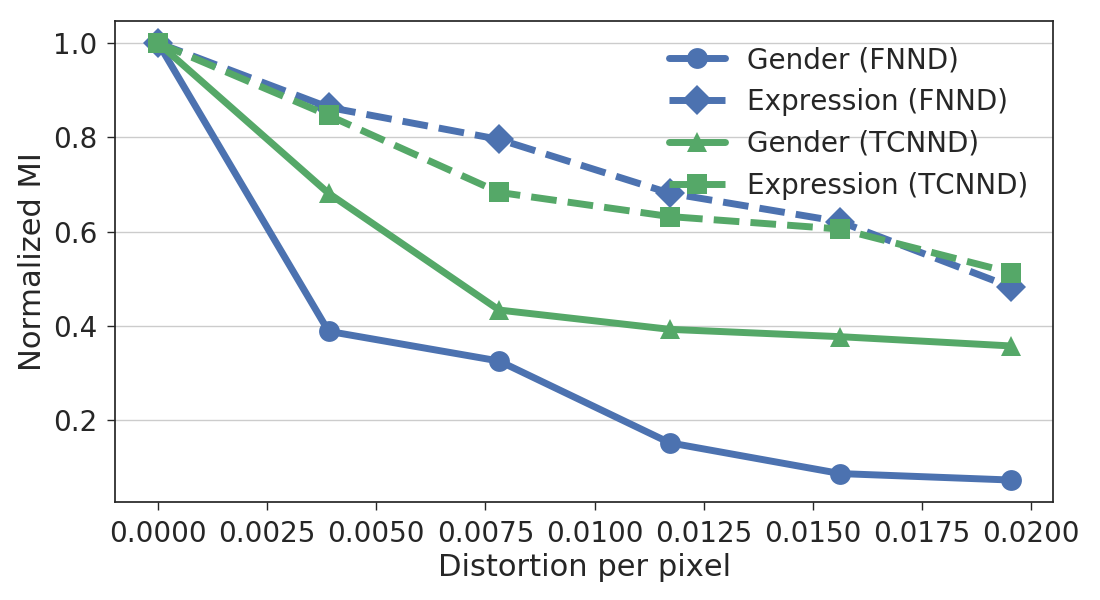}
		\caption{Normalized mutual information estimation}
		\label{fig:genki:mi_normed}
	\end{subfigure}
	\caption{The tradeoff between classification accuracy and mutual information estimation for the GENKI dataset.}%
	\label{fig:privacyutility_mi}
\end{figure}

\begin{table*}[]
\centering
\small
\begin{tabular}{|c|c|c|c|c|c|c|c|}
\hline
Distortion $D$                                                                                & 1       & 2       & 3       & 4       & 5       & 100    & 1000   \\ \hline
\begin{tabular}[c]{@{}c@{}}Laplace Mechanism $\epsilon$\end{tabular}                        & 5792.61 & 4096    & 3344.36 & 2896.31 & 2590.53 & 579.26 & 183.17 \\ \hline
\begin{tabular}[c]{@{}c@{}}Gaussian Mechanism ($\delta = 10^{-6}$)  $\epsilon$\end{tabular} & 1918.24 & 1354.08 & 1107.57 & 959.18  & 857.76  & 191.82 & 60.66  \\ \hline
\end{tabular}
\caption{Differential privacy risk for different distortion values.}
\label{tb:genki}
\end{table*}

\begin{table*}
\centering
\begin{tabular}{|c|c|c|c|c|c|c|c|c|}
\hline
{\begin{tabular}[c]{@{}c@{}} Expression \end{tabular}} & \multicolumn{2}{c|}{Original Data} & \multicolumn{2}{c|}{D = 1} & \multicolumn{2}{c|}{D = 3} & \multicolumn{2}{c|}{D = 5} \\ \cline{2-9} 
Classification & Male            & Female           & Male        & Female       & Male        & Female       & Male        & Female       \\ \hline
False Positive Rate & 0.04                & 0.14                  & 0.1        & 0.18         & 0.18        & 0.16          & 0.16            &0.14              \\ \hline
False Negative Rate & 0.16                & 0.02                 & 0.2        & 0.08         & 0.26        & 0.12         & 0.24            &0.24              \\ \hline
\end{tabular}

\caption{Error rates for expression classification using representation learned by FNND for the GENKI dataset.}
\label{tb:errorFNNE}
\vspace{0.1in}
\begin{tabular}{|c|c|c|c|c|c|c|c|c|}
\hline
{\begin{tabular}[c]{@{}c@{}} Expression \end{tabular}} & \multicolumn{2}{c|}{Original Data} & \multicolumn{2}{c|}{D = 1} & \multicolumn{2}{c|}{D = 3} & \multicolumn{2}{c|}{D = 5} \\ \cline{2-9} 
Classification & Male            & Female           & Male        & Female       & Male        & Female       & Male        & Female       \\ \hline
False Positive Rate&  0.04               & 0.14                 & 0.04        & 0.16         & 0.06        & 0.12          &0.08             & 0.16             \\ \hline
False Negative Rate&  0.16                & 0.02                 & 0.2        & 0.08         & 0.2        & 0.14         &0.18             &0.16              \\ \hline
\end{tabular}
\caption{Error rates for expression classification using representation learned by TCNND for the GENKI dataset.}
\label{tb:errorTCNNE}
\end{table*}
\noindent{\textbf{Evaluating adversarial performance via mutual information estimation}: Our FUR framework offers a scalable way to find a (local) equilibrium in the constrained min-max optimization for certain adversarial attacks (e.g. inference attacks on $S$ using a neural network). Yet the privatized data, through our approach, should be immune to any general attacks and should ultimately achieve the goal of decreasing the correlation between the $X_r$ and $S$. To this end, we use the estimated MI, using the $k$-nearest neighbor method as detailed in Section \ref{sec:GMM-results}, to verify that our framework protects $S$.  }

{One noteworthy difficulty is that $X$, and therefore, $X_r$ are usually high dimensional objects (each image in the GENKI dataset has $256$ dimensions), so it is almost impossible to calculate the empirical entropy based on raw data due to the sample complexity.  Thus, we train a neural network that classifies $S$ from the learned data representation to reduce the dimension of the data. We choose the layer before the softmax outputs of the adversary (denoted by $\hat{X}_g$) to be the feature embedding that has a much lower dimension than the original $X_r$ which still captures the information about $S$. We use this $\hat{X}_g$ as a surrogate for $X_r$ in \eqref{eq:k-nn-entropy-est} to first estimate $\hat{H}(\hat{X}_g)$ and then compute $\hat{I}(\hat{X}_g; S)$ as the approximate MI between $X_r$ and $S$.  
We similarly extract an $\hat{X}_f$ by training a neural network that now classifies $Y$; we then compute $\hat{H}(\hat{X}_f)$ from which we obtain $\hat{I}(\hat{X}_f; Y)$ as the approximate MI between $X_r$ and $Y$.  
The details of the common neural network architecture we use to extract $\hat{X}_f$ and $\hat{X}_g$ can be found in the supplement (see Section SII-C).}



\noindent{\textbf{Utility-fairness tradeoffs}: To evaluate the value of the representation generated by the FUR  framework for the GENKI dataset, we consider the task of classifying the facial expression (non-sensitive feature $Y$) as smiling or non-smiling (i.e., the task that the GENKI dataset was intended for). To this end, we train another CNN (see Fig. S4 in the supplement for architecture details) to perform facial expression classification on datasets processed by different decorrelation schemes. The trained model is then tested on the original test data. In Fig.~\ref{fig:genkiprivacy_util}, we observe that the expression classification accuracy decreases gradually as the distortion increases. However, even for a large distortion value (5 per image), the expression classification accuracy only decreases by $10\%$. To make meaningful comparisons using MI, we normalize $\hat{I}(\hat{X}_f; Y)$ by $\hat{I}(X_f;Y)$ where $X_f$ is the low-dimensional representation of $X$ (=$X_r$ for $D=0$); we similarly normalize $\hat{I}(\hat{X}_g; S)$ by $\hat{I}(X_g;S)$ where $X_g$ is defined similarly. As shown in Fig. \ref{fig:genki:mi_normed}, the estimated normalized MI $\hat{I}(\hat{X}_f; Y)/\hat{I}(X_f;Y)$ decreases at a much slower rate than $\hat{I}(\hat{X}_g; S)/\hat{I}(X_g;S)$ as the distortion increases thus verifying that $X_r$ preserves relatively more information about $Y$ than it does about $S$ at every distortion level.}

\begin{figure*}[t]
\centering
\small
	\includegraphics[width=1.65\columnwidth]{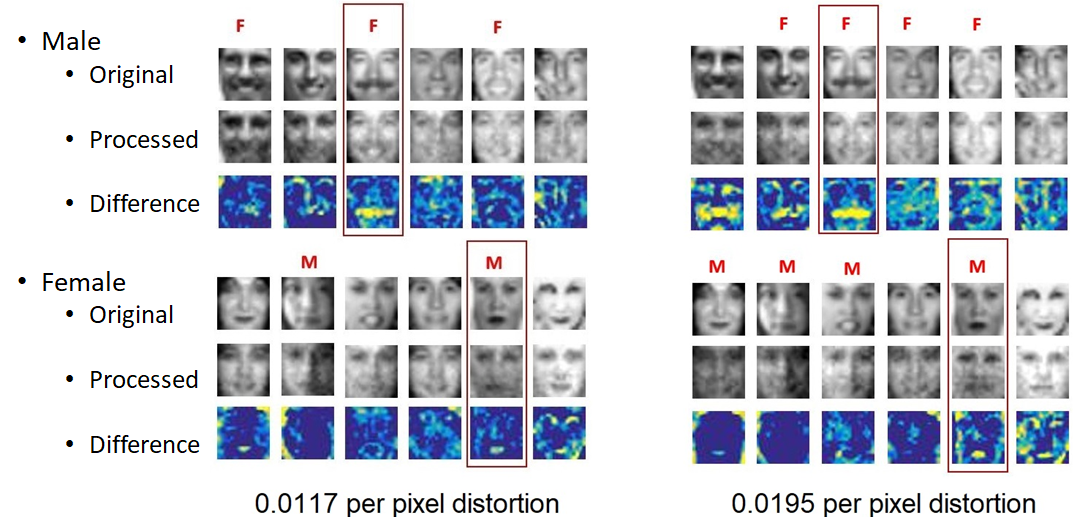}
	\caption{Perturbed images with different per pixel distortion using FNND.}
	\label{fig:privatizedimages}
\end{figure*}

{Tables \ref{tb:errorFNNE} and \ref{tb:errorTCNNE} present different error rates for the facial expression classifiers trained using data representations created by different decorrelator architectures. For the GENKI dataset, a smiling expression is considered as the positive label for the expression classification task. We observe that as distortion increases, the difference across the two sensitive groups (male vs. female) for each error rate decreases. This implies the classifier's decision is less biased with respect to $S$ when trained using $X_r$. When $D = 5$, the differences are quite small. In particular, the FNND architecture performs better in enforcing fairness but suffers from a slightly higher error rate relative to TCNND.}
The images processed by FNND are shown in Fig. \ref{fig:privatizedimages}. As highlighted in the figure, the dominant features that the decorrelator changes are those that capture gender, namely eyes, nose, mouth, beard, and hair.

\vspace{-0.15in}
{
\subsection{Illustration of Results for HAR Dataset}
\label{sec:HAR}
For the HAR dataset, we recall that the $X$ features are motion sensor data from 30 subjects performing six different activities; the goal is to classify activity (intended task $Y$) without revealing subject identity (sensitive $S$).
In Fig.~\ref{fig:haraccuracy}, for different values of distortion, we illustrate the accuracy in classifying activity against that of identification. We see that classification accuracy of identity decreases 
as the distortion increases. In fact, when distortion is small ($D=2$), the identity accuracy is down to $27\%$. If we increase the distortion to $8$, the identity accuracy further decreases to $3.8\%$. 
However, we note that for an even large distortion value ($D=8$), the activity classification accuracy only decreases by at most $18\%$. Finally, in Fig. \ref{fig:harMI}, we demonstrate that the estimated normalized mutual information $\hat{I}(\hat{X}_f; Y)/\hat{I}(X_f;Y)$ decreases at a much slower rate than $\hat{I}(\hat{X}_g; S)/\hat{I}(X_g;S)$ as the distortion increases, thereby assuring that the adversarial model chosen assures censoring of identity without compromising accuracy of the intended task from $X_r$. Details on the architecture for the FUR model, the classifier, and for obtaining the representations used to estimate MI can be found in the supplement. 

\begin{figure}[h]
	\centering
	\small
	\begin{subfigure}[t]{0.45\textwidth}
		\includegraphics[width=1\columnwidth]{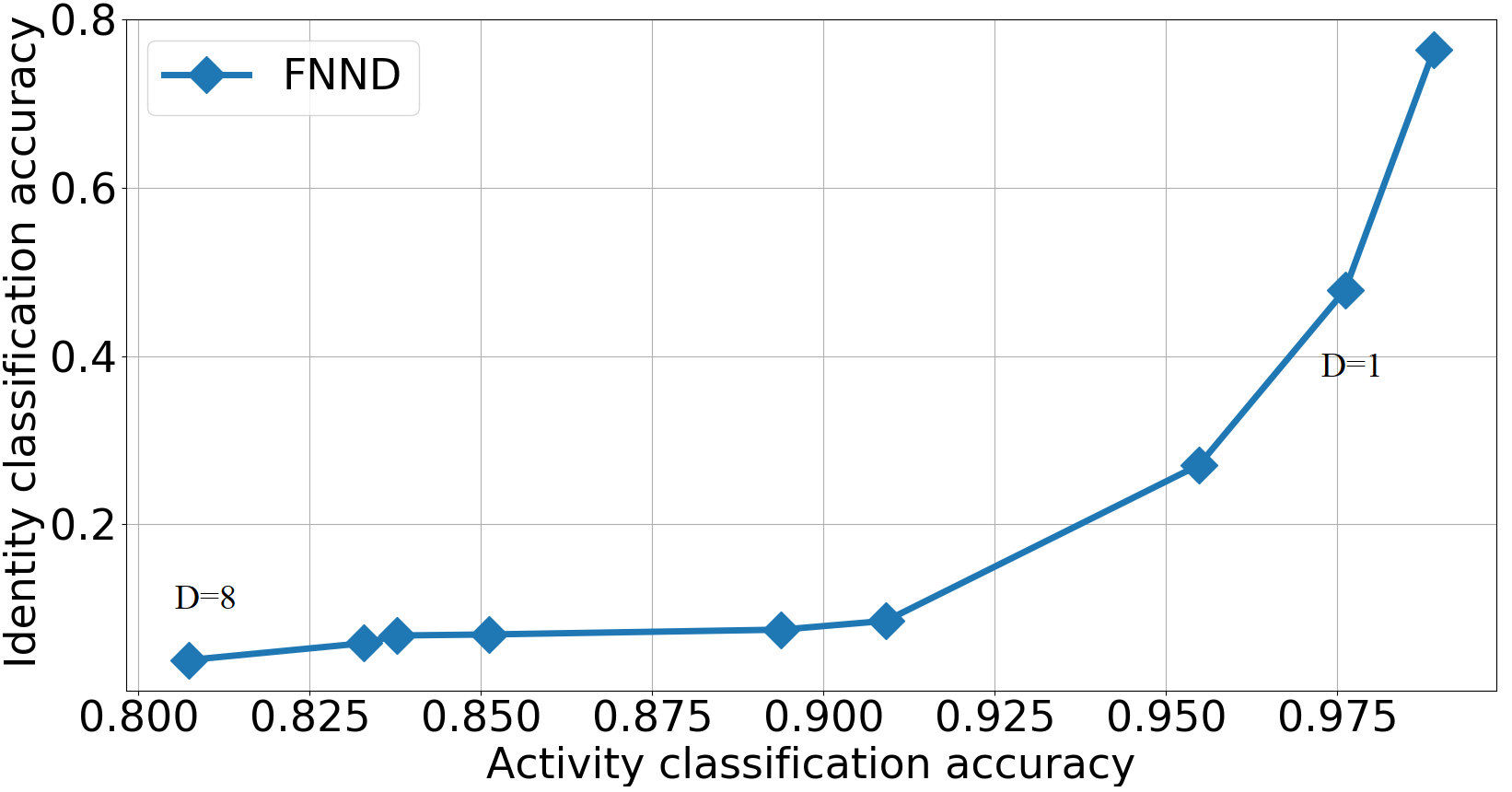}
		\caption{Identity vs. activity classification accuracy}
		\label{fig:haraccuracy}
	\end{subfigure}
	\begin{subfigure}[t]{0.46\textwidth}
		\includegraphics[width=1\columnwidth]{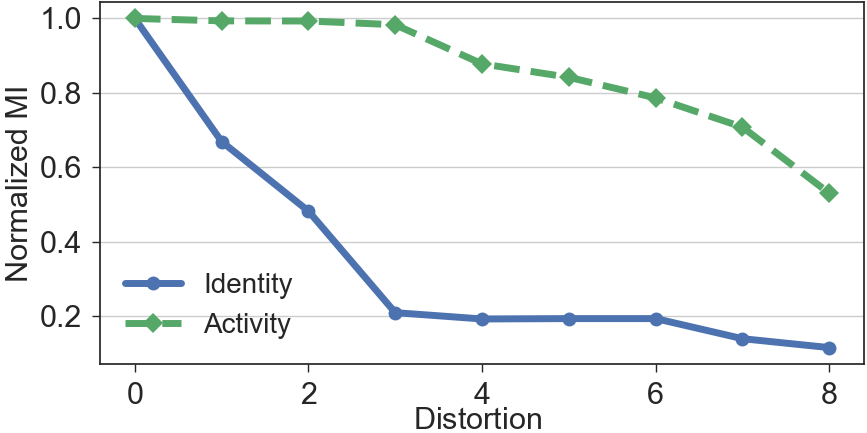}
		\caption{Normalized mutual information estimation}
		\label{fig:harMI}
	\end{subfigure}
	\caption{The tradeoff between classification accuracy and mutual information estimation for the HAR dataset.}%
	\label{fig:harprivacy}
	\vspace{-0.1in}
\end{figure}

\vspace{-0.1in}
\section{Conclusion}
\label{sec:conclusion}
We have introduced an adversarial learning framework with verifiable guarantees for learning generative models that can create censored and fair universal representations for datasets with known sensitive features. The novelty of our approach is in producing representations that are fair with respect to the sensitive features for any \textit{a priori} unknown downstream learning task. We have shown that our FUR framework allows the data holder to learn the fair encoding scheme (a randomized mapping that decorrelates the {sensitive and non-sensitive} features) directly from the dataset without requiring access to dataset statistics. 
One of our key results is that for appropriately chosen loss functions, the minimax game generating FURs can provide guarantees against strong information-theoretic adversaries, such as MAP (0-1 loss) and MI (log-loss) adversaries. We have also shown that our framework also allows approaching (ideal) demographic parity fairness using a tunable class of $\alpha$-loss functions (which includes both log-loss and 0-1 loss) that capture a range of adversarial actions. For the setting with a known classification task, we have also shown that our FUR framework can be modified to approximate either DemP or EO fairness measures. 

Finally, we have also validated the performance of the FUR framework on both synthetic and publicly available real datasets, including Gaussian mixture models, images, and datasets involving a mix of categorical and continuous features. Our results have allowed us to visually highlight three key results: (a) the tradeoff between representation fidelity and censoring guarantees (via accuracy in adversarially learning sensitive features); (b) the tradeoff between the adversarial accuracy in learning the sensitive features and the accuracy of multiple downstream tasks learned from the censored representation for a variety of datasets; and (c) the tradeoff between accuracy for a downstream learning task and the DemP or EO guarantees achieved by a predictor that is learned using a DemP fair representation. Result (c) further suggests that, for some datasets, DemP FURs do not adversely affect the downstream EO guarantees despite lack of access to the task labels $Y$, where the latter has been highlighted as a limitation of DemP fair predictors~\cite{hardt2016equality}. However, more work is needed to understand the conditions under which DemP FURs suffice to achieve meaningful EO guarantees. {Such an exploration can also clarify if the definition of $\Delta_{\text{DemP}}$ as the worst case over all task output values $y\in\mathcal{Y}$ and sensitive values $s \in\mathcal{S}$ is too strong, especially for settings where $\mathcal{Y}$ and/or $\mathcal{S}$ are large, as our age prediction results suggest for the UTKFace dataset.}

Going beyond this work, there are several questions that can be addressed. 
An immediate one is to explore the robustness of using $\alpha$-loss (for $\alpha\ne 1$) in ensuring censoring and fairness for highly imbalanced datasets. {Yet another is to explore the usefulness of FURs for unsupervised tasks building on recent work in \cite{lahoti2019ifair}.}
More broadly, it will be interesting to investigate the robustness and convergence guarantees of the generative encoder learned in a data-driven fashion. 
\section{Acknowledgement}
We would also like to thank Prof. Ram Rajagopal and Xiao Chen at Stanford for their help with estimating mutual information for the GENKI and HAR datasets and would like to thank Mit Patel and Harshit Laddha at ASU for their help with the experiments on the UTKFace dataset.
\vspace{-0.2in}
\appendix
\if \extended 1
\subsection{Proof of Theorem \ref{thm:DemP_broadcasting}}
\label{Proof:thm:DemP_broadcasting}
A demographically fair encoder $g(X,S)$ ensures that $X_r\perp S$,
i.e., the mutual information $I(S;X_r)=0$. Further, the downstream learning algorithm acts only on $X_r$ to predict $\hat{Y}$; thus, $(S,X)-X_r-\hat{Y}$ form a Markov chain.
From the data processing inequality and non-negativity of mutual information, we have $0\leq I(S;\hat{Y})\leq I(S;X_r)=0$, i.e.,
$S$ is independent of $\hat{Y}$; thus, from Definition \ref{Def:ThreeFairnessMeasures}, $\hat{Y}$ satisfies DemP w.r.t. $S$.
Finally, from the chain rule and non-negativity of mutual information, we have $I(X_r;S_t)=0$ for any $S_t \subset S$, i.e., $\hat{Y}$ satisfies DemP w.r.t. any subset of sensitive features $S_t$.

\vspace{-0.2in}
\section{Proofs for Results in Section \ref{sec:model}}
\label{pf:sec:model}

\subsection{Proofs of Proposition \ref{thm:alpha}, Results in Table  \ref{table:gaploss}, and Theorem \ref{thm:DP}}\label{proof:table:gaploss}
In the FUR framework, different loss functions and decision rules lead to different adversarial models. We now prove the optimal adversarial rule for different loss functions in Table \ref{table:gaploss}. 

\noindent\textbf{Hard Decision Rules. } When the adversary adopts a hard decision rule, $h(g(X))$ is an estimate of $S$. Under this setting, we can choose $\ell(h(g(X)), S)$ in a variety of ways. For instance, if $S$ is continuous, the adversary can attempt to minimize the difference between the estimated and true sensitive variables using a squared loss ($\ell_2$ loss) as
\begin{align}
\ell(h(g(X)),S)= (h(g(X)) - S)^2.
\end{align}
One can verify that the optimal $h^* = \mathbb{E}[S|g(X)]$, the conditional mean of $S$ given $g(X)$. Then, \eqref{eq:generalopt} simplifies to
$$\min_{g(\cdot)} - \text{mmse}(S|g(X)) = - \max_{g(\cdot)} \text{mmse}(S|g(X)),$$
subject to the distortion constraint and mmse is the minimum mean square error (MMSE). 
Thus, for the $\ell_2$ loss, the FUR framework provides censoring guarantees against an MMSE adversary. On the other hand, when $S$ is discrete (e.g., age, gender, etc.), the adversary can maximize its classification accuracy by considering the $0$-$1$ loss \cite{nguyen2013algorithms} given by
\begin{align}
\label{eq:0-1loss}
\ell(h(g(X)),S)=
\left\{
\begin{array}{ll}
0  & \mbox{if } h(g(X))=S \\
1  & \mbox{otherwise. }
\end{array}
\right.
\end{align}
One can verify that the adversary's optimal strategy is the maximum \textit{a posteriori} probability (MAP) rule: $h^* = \argmax_{s \in \mathcal{S}} P(s|g(X))$, with ties broken uniformly at random. For this MAP rule, the objective in \eqref{eq:generalopt} reduces to
\begin{equation*}
\label{eq:mapadversaryaccuracy}
 \min\limits_{g(\cdot)} - (1 - \max\limits_{s \in \mathcal{S}} P(s, g(X))) = \min\limits _{g(\cdot)} \max_{s \in \mathcal{S}} P(s, g(X))  - 1,
\end{equation*}
i.e., under $0$-$1$ loss, the FUR formulation provides censoring guarantees against a MAP adversary.\\
\noindent\textbf{Soft Decision Rules. } Instead of a \textit{hard decision}, we can also consider a broader class of \textit{soft decision} rules where $h(g(X))$ is a distribution over $\mathcal{S}$; i.e., $h(g(X)) = P_h(s|g(X))$ for $s \in \mathcal{S}$. We first consider $\log$-loss given by
\begin{align}
\label{eq:infologloss}
\ell(h(g(X)),s) = \log \frac{1}{P_h(s|g(X))}.
\end{align}
In this case, the objective of the adversary simplifies to
$$ \max\limits_{h(\cdot)}  -\mathbb{E}\left[\log \frac{1}{P_h(s|g(X))}\right] = - H(S|g(X)),$$
where $H(X|Y)$ is the conditional entropy of $X$ given $Y$ and, given the true posterior $P(s|g(X))$, $P^*_h(s|g(X)) = P(s|g(X))$. 
Then, the objective in \eqref{eq:generalopt} reduces to
\begin{align}
\label{eq:infologloss}
\min_{g(\cdot)}  - H(S|g(X)) = \min_{g(\cdot)} I(g(X);S) - H(S),
\end{align}
i.e., under $\log$-loss, the FUR formulation is equivalent to using mutual information as the censoring (or privacy) metric.\\ 
\noindent\textbf{Tunable $\alpha$-loss family.} For $\ell(h(g(x)),s)$ chosen as $\alpha$-loss in \eqref{eq:alpha_loss}, we have for $\alpha \in (0, \infty)$,
\begin{align} 
 & \max_{h(\cdot)}  -\mathbb{E}\bigg[\frac{\alpha}{\alpha  - 1} \left( 1- P_h(S|g(X))^{1 - \frac{1}{\alpha}}\right)\bigg]  =\nonumber \\
 & \frac{\alpha}{\alpha-1}\bigg (\sum_x P(g(x)) \max_{h(\cdot)} \sum_{s} P(s|g(x))  P_h(s|g(x))^{1 - \frac{1}{\alpha}} - 1 \bigg),
 \label{eq:max}
\end{align}
with limits taken appropriately for $\alpha=0,1,\text{ and } \infty$. For each $x \in \mathcal{X}$ with $P(g(x)) > 0$, the maximization in \eqref{eq:max} can be written as
\begin{equation}
\begin{aligned}
 \max_{h(\cdot)} \quad & \frac{\alpha}{\alpha -1 } \sum_{s \in \mathcal{S}} P(s|g(x)) P_h(s|g(x))^{1 - \frac{1}{\alpha}}\\
\textrm{s.t.} \quad & \sum_{s \in \mathcal{S}} P_h(s|g(x)) = 1, \\
  & P_h(s|g(x))\geq0, \quad \text{for all } s \in  \mathcal{S}.
\end{aligned}
\label{eq:opt_prob}
\end{equation}
For $\alpha \in [0, \infty]$, the optimization in \eqref{eq:opt_prob} is convex in $P_h(s|g(x))$. Hence, we can use Karush Kuhn-Tucker (KKT) conditions to solve for the optimal $h^*$, ignoring the inequality constraint at first and then checking that the resulting $h^*$ satisfies the constraint. Consider the following Lagrangian:
\begin{equation}
\begin{aligned}
    L(h(\cdot),\lambda) = \frac{\alpha}{\alpha -1 }\sum_{s \in \mathcal{S}} & P(s|g(x)) P_h(s|g(x))^{1 - \frac{1}{\alpha}} \\
    + & \lambda \bigg(\sum_{s \in \mathcal{S}}  P_h(s|g(x)) - 1\bigg).
\end{aligned}
\end{equation}
Noting that $P_h(s|g(x))$ is a vector of probabilities over all $S$ for a given $g(x)$, for every $s$, at the optimal $h^*(g(x))=P_h^*(s|g(x))$, we obtain the following KKT conditions:
\begin{align*}
    \nabla_{h(\cdot)} L(P_h^*(s|g(x)),\lambda^*) = 0, \quad \text{i.e.}
\end{align*}
\begin{equation}
    P(s|g(x)) P_h^*(s|g(x))^{- \frac{1}{\alpha}} + \lambda^* = 0,
\label{eq:lagrangian_opt}
\end{equation}
such that
\begin{equation}
    \sum_{s \in \mathcal{S}} P_h^*(s|g(x)) = 1.
\label{eq:feasibility}
\end{equation}
Solving \eqref{eq:lagrangian_opt} for $h^*$, we get
\begin{align*}
    P_h^*(s|g(x)) = \lambda^\prime P(s|g(x))^\alpha,
\end{align*}
where $\lambda^\prime = \big(\frac{-1}{ \lambda^*}\big)^\alpha$. Enforcing \eqref{eq:feasibility}, we find that $\lambda^\prime = 1/\sum_{s \in \mathcal{S}} P(s|g(x))^\alpha$, and therefore, the optimal $h^*$ is a `$\alpha$-tilted' conditional distribution
\begin{equation} \label{eq:opt-h-alpha-loss}
P^*_h(s|g(X)) = {P(s|g(X))^\alpha}/{\sum_{s'\in\mathcal{S}} P(s'|g(X))^\alpha}.
\end{equation}
Thus, for the optimal adversarial strategy $P^*_h(s|g(X))$ in \eqref{eq:opt-h-alpha-loss}, the objective simplifies as:
\begin{align} 
 \max_{h(\cdot)} &  -\mathbb{E}\bigg[\frac{\alpha}{\alpha  - 1} \left( 1- P_h(S|g(X))^{1 - \frac{1}{\alpha}}\right)\bigg] \nonumber \\
 & = \frac{\alpha}{1-\alpha}\left(1-\exp\left(\frac{1-\alpha}{\alpha}H_{\alpha}^{\text{A}}(S|g(X))\right)\right) 
\end{align}
where  $H^{\text{A}}_{\alpha}(U|V)$ for two discrete random variables $(U,V)\sim P_{U,V}(u,x)$
is the Arimoto conditional entropy of order $\alpha$ defined as
\begin{equation}
\label{eq:arimoto-cond-entropy}
    H^{\text{A}}_{\alpha}(U|V)\triangleq \frac{1}{(1-\alpha)}\log \left(\sum_{u,v}P_{U,V}^\alpha(u,v)\right).
\end{equation}
The resulting minimax game reduces to
\begin{equation}
\label{eq:alpha_minmax}
\min\limits_{g(\cdot)}  I^{\text{A}}_{\alpha}(S;g(X)) - H_\alpha(S),
\end{equation}
where $H_\alpha(S)$ is the R\'enyi entropy of order $\alpha$ defined as $H^{\text{A}}_{\alpha}(S)\triangleq 1/(1-\alpha)\log(\sum_{s}P_{S}^\alpha(s))$ and we have used the fact that for two discrete random variables $(U,V)\sim P_{U,V}(u,x)$, the Arimoto MI is $I_\alpha(U;V)=H_\alpha(U)-H_\alpha(U|V)$. As a side-note, we observe that generalizing the Renyi entropy to a conditional form leads to many different formulations and the Arimoto conditional entropy in \eqref{eq:arimoto-cond-entropy} is one such form introduced by Arimoto \cite{arimoto1977information}. 

For $\alpha=1$, \eqref{eq:alpha_minmax} simplifies to \eqref{eq:infologloss} since the Arimoto conditional entropy (resp. MI) simplifies to the Shannon conditional entropy (resp. MI) \cite{alphaMI_verdu}. 
From the non-negativity of Arimoto MI $I_\alpha^\text{A}(S;g(X))$, we also have $H_{\alpha}^{\text{A}}(S|g(X))\leq H_{\alpha}(S)$ \cite{alphaMI_verdu}
with equality if and only if $g(X)$ is independent of $S$. Thus, the minimization in \eqref{eq:alpha_minmax} serves as a proxy for generating DemP fair representations with perfect DemP achieved when $I_\alpha^\text{A}=0$.

Near perfect DemP is achievable for very large distortions as this allows $X_r$ to be increasingly decoupled from $S$ and $X$. Indeed, as the bound $D$ in \eqref{eq:generalopt} increases, from Proposition \ref{Thm:GAP_censoring}, the FUR formulation in \eqref{eq:generalopt} will \emph{approach} ideal DemP by achieving  $I_{\alpha}^{\text{A}}(S|g(X))\approx 0$, thus proving Theorem \ref{thm:DP}.

\vspace{-0.2in}
\subsection{Proof of Theorem \ref{prop:EO}}\label{Proof:prop:EO}
The optimization in \eqref{eq:Opt-fairClassifier-EO} differs from that in \eqref{eq:generalopt} in including the target $Y$ in learning both $g$ and $h^{\star}$. In fact, for a fixed ${g}$, one can verify that the optimal adversarial strategy in \eqref{eq:Opt-fairClassifier-EO} is the same as \eqref{eq:opt-h-alpha-loss} with an additional conditioning on $Y$. Thus, for $\ell=\ell_\alpha$ in \eqref{eq:alpha_loss}, the optimization in \eqref{eq:Opt-fairClassifier-EO} reduces to 
\begin{equation}
\label{eq:Opt-fairClassifier-EO-InPf}
\min_{{g}(\cdot)}  I_{\alpha}^{\text{A}}(S;{g}(\cdot)|Y),
\:
\textrm{s.t.} \: \mathbb{E}[\ell({g}(\cdot),Y)]\le \epsilon.
\end{equation}

Since $I_{\alpha}^{\text{A}}(S;\tilde{g}(\cdot)|Y)\geq 0$
with equality if and only if ${g}(\cdot) \perp S|Y$, minimizing \eqref{eq:Opt-fairClassifier-EO} with $\alpha$-loss achieves EO. In practice, as the performance constraint $\epsilon$ in \eqref{eq:Opt-fairClassifier-EO-InPf} increases, the FUR formulation in \eqref{eq:Opt-fairClassifier-EO-InPf} will approach perfect EO of ${g}(\cdot)$ w.r.t. $Y$ and $S$ by enforcing $H_{\alpha}^{\text{A}}(S|{g}(S, X),Y)= H_{\alpha}^{\text{A}}(S|Y)$.

\vspace{-0.15in}
\fi
\subsection{Alternate Minimax Algorithm}
\label{sec:alternateminimax}
Algorithm \ref{alg:euclid} details the steps used to learn the FUR model in a data-driven manner. To incorporate the distortion constraint, we use the \textit{penalty method} \cite{lillo1993solving} 
to replace the constrained optimization problem by a series of unconstrained problems. 
The unconstrained optimization problem is formed by adding a penalty to the objective function as a product of a parameter $\rho_t$ and an appropriate measure of violation of the constraint.  
We start with a large value of $\rho_t$ to enforce distortion from the outset and decrease $\rho_t$ in exponential steps with respect to the number of training epochs. Such a decrease allows enforcing a smaller penalty when the model is closer to convergence. Finally, we also vary the learning rate $\eta_t$ over training epochs as follows: we pick a small value of $\eta_t$ at the beginning and compare the relative values of the adversarial loss and the average distortion. We adjust the initial $\eta_t$ so that the adversarial loss and the distortion penalty values are on a similar scale in the first few epochs during training. When the algorithm terminates, we check the average distortion and manually fine tune the initial $\eta_t$ and the update rule to make sure the distortion is within bounds after termination.

\begin{algorithm}[]
	\caption{Alternating minimax FUR algorithm}\label{alg:euclid}
	\begin{algorithmic}
		\State \textit{Input:} dataset $\mathcal{D}$, distortion parameter $D$, \# of decorrelator iterations $T$, \# of adversary iterations $J$ for each round of decorrelator update, minibatch size $M$
		\State \textit{Output:} Optimal generative decorrelator parameter $\theta_{p}$
		\Procedure{Alernate Minimax}{$\mathcal{D}, D, T, J, M$}
		\State Initialize decorrelator parameter $\theta^1_{p}$, adversary parameter $\theta^1_{a}$, and step size $\eta_1$
		\For{$t=1,...,T$}
		\State Random minibatch of $M$ datapoints $\{x_{(1)},...,x_{(M)}\}$ drawn from full dataset
		\State Generate $\{\hat{x}_{(1)},...,\hat{x}_{(M)}\}$ via $\hat{x}_{(i)}=g(x_{(i)};\theta^t_{p})$ 
		\State Apply update rule for step size $\eta_t$
		\State Set $\omega^1_a$ = $\theta^t_a$
		\For{$j=1,...,J$}
		\State Update the adversary parameter $\theta^{t+1}_a$ by stochastic gradient ascent for epoch $j$ 
		{$$\quad \omega^{j+1}_a=\omega^{j}_a+\eta_t\nabla_{\omega^{j}_a} \frac{1}{M}\sum\limits_{i=1}^{M}-\ell(h(\hat{x}_{(i)};\omega^{j}_{a}),s_{(i)}), \quad\eta_t>0$$}
		\EndFor
		\State Set $\theta^{t + 1}_{a}$ = $\omega^{J + 1}_a$
		\State Compute the descent direction $\nabla_{\theta^{t}_{p}} L_m(\theta^{t}_{p}, \theta^{t+1}_a)$, where $L_m(\theta^{t}_{p},\theta^{t+1}_a)$ is defined in \eqref{eq:gap-CE-datadriven} for $n=m$
%
		\State Perform line search along $ \nabla_{\theta^{t}_{p}} L_m(\theta^{t}_{p}, \theta^{t+1}_a)$ and, for $\ell(\theta^{t}_{p},\theta^{t+1}_a)$ set as the objective in \eqref{eq:penaltymethodsupp} for $n=m$, update
		$$\theta^{t+1}_{p}= \theta^{t}_{p}-\eta_t \nabla_{\theta^{t}_{p}} \ell(\theta^{t}_{p}, \theta^{t+1}_a)$$
		\EndFor
		\State \textbf{return} $\theta^{T+1}_p$
		\EndProcedure
	\end{algorithmic}
\end{algorithm}

\vspace{-0.17in}
\subsection{Proof of Theorem \ref{thm:PDI}}
\label{proof:PDI}

Since $\mathbb{E}_{{X},\hat{{X}}}[ d(\hat{{X}}, {X})] =\mathbb{E}_{{X},\hat{{X}}}\lVert {X}-\hat{{X}}\rVert^2=\mathbb{E}\lVert{Z}+{\beta}\rVert^2= \lVert{\beta}\rVert^2+tr(\Sigma_p)$, the distortion constraint implies that $\lVert{\beta}\rVert^2+tr(\Sigma_p) \leq D$. Let us consider $\hat{X} = X + Z + \beta$, where $\beta\in\mathbb{R}^m$ and $\Sigma_p$ is a diagonal covariance whose diagonal entries are given by $\{\sigma^2_{p_{1}},...,\sigma^2_{p_{m}}\}$. Given the MAP adversary's optimal inference accuracy in \eqref{eq:gaussianscheme0}, the objective of the decorrelator is
\begin{align}
\label{eq:gaussianscheme0opt}
\min\limits_{\beta,\Sigma_p} {P^{\text{(G)}}_{\text{d}}}, \: \textrm{s.t.} \: \lVert{\beta}\rVert^2+tr(\Sigma_p) \leq D.
\end{align}

Define $\frac{1}{\gamma}\ln{\frac{1-{q}}{{q}}}=\eta$.
After some algebra, the gradient of ${P^{\text{(G)}}_{\text{d}}}$ \textit{w.r.t.} $\gamma$ is given by
\begin{equation*} 
\begin{aligned}
\frac{\partial {P^{\text{(G)}}_{\text{d}}}}{\partial \gamma}
=& \frac{1}{2\sqrt{2\pi}}\left(qe^{-\frac{\left(\eta -  { \frac{\gamma}{2} } \right) ^2}{2}}+(1-q)e^{-\frac{\left(\eta +  { \frac{\gamma}{2} } \right) ^2}{2}}\right),
\end{aligned}
\end{equation*}
which is always positive. Thus, ${P^{\text{(G)}}_{\text{d}}}$ is monotonically increasing in $\gamma$. As a result, \eqref{eq:gaussianscheme0opt} is equivalent to
\begin{align}
\label{eq:gaussianscheme0opt_2}
\min\limits_{{\beta},\sigma_{p_{1}}^2,...,\sigma_{p_{m}}^2} \quad & \sum_{i=1}^{m} \frac{\mu_i^2}{\sigma_i^2+\sigma_{p_{i}}^2},\\\nonumber
 \textrm{s.t.}\quad & \lVert{\beta}\rVert^2+tr(\Sigma_p) \leq D\\\nonumber
 & \sigma_{p_{i}}^2\ge 0 \quad\forall i\in\{1,2,...m\}.
\end{align}
The optimization in \eqref{eq:gaussianscheme0opt_2} is analogous to the well-studied rate-distortion problem for independent Gaussian sources and the optimal solution given by reverse water-filling \cite[Chap. 10.3.3]{cover2012elements}. Using similar techniques, we obtain ${\sigma^*_{p_{i}}}^2=\max\{\lvert\mu_i\rvert/\sqrt{\lambda^*_0}-\sigma_i^2,0\}=\left({\lvert\mu_i\rvert}/{\sqrt{\lambda^*_0}}-\sigma_i^2\right)^{+}$ with $\sum_{i=1}^{m}{\sigma^*_{p_{i}}}^2=D$ leading to the results in the theorem. 

 \vspace{-0.1in}
\bibliographystyle{IEEEtran}
\bibliography{Bib-Files}

\end{document}


\title{Generating Fair Universal Representations using Adversarial Models -- Supplementary Material}

\maketitle
\graphicspath{{Figures/}}


\newpage

\section{Overview}
This document contains the architectural details of the UCI Adult, UTKFace, GENKI, and HAR dataset experiments presented in the main paper, as well as the details on computing differential censoring risk for Laplace and Gaussian noise for the GENKI dataset. 
\begin{figure}[ht]
\centering
\includegraphics[width=1\columnwidth]{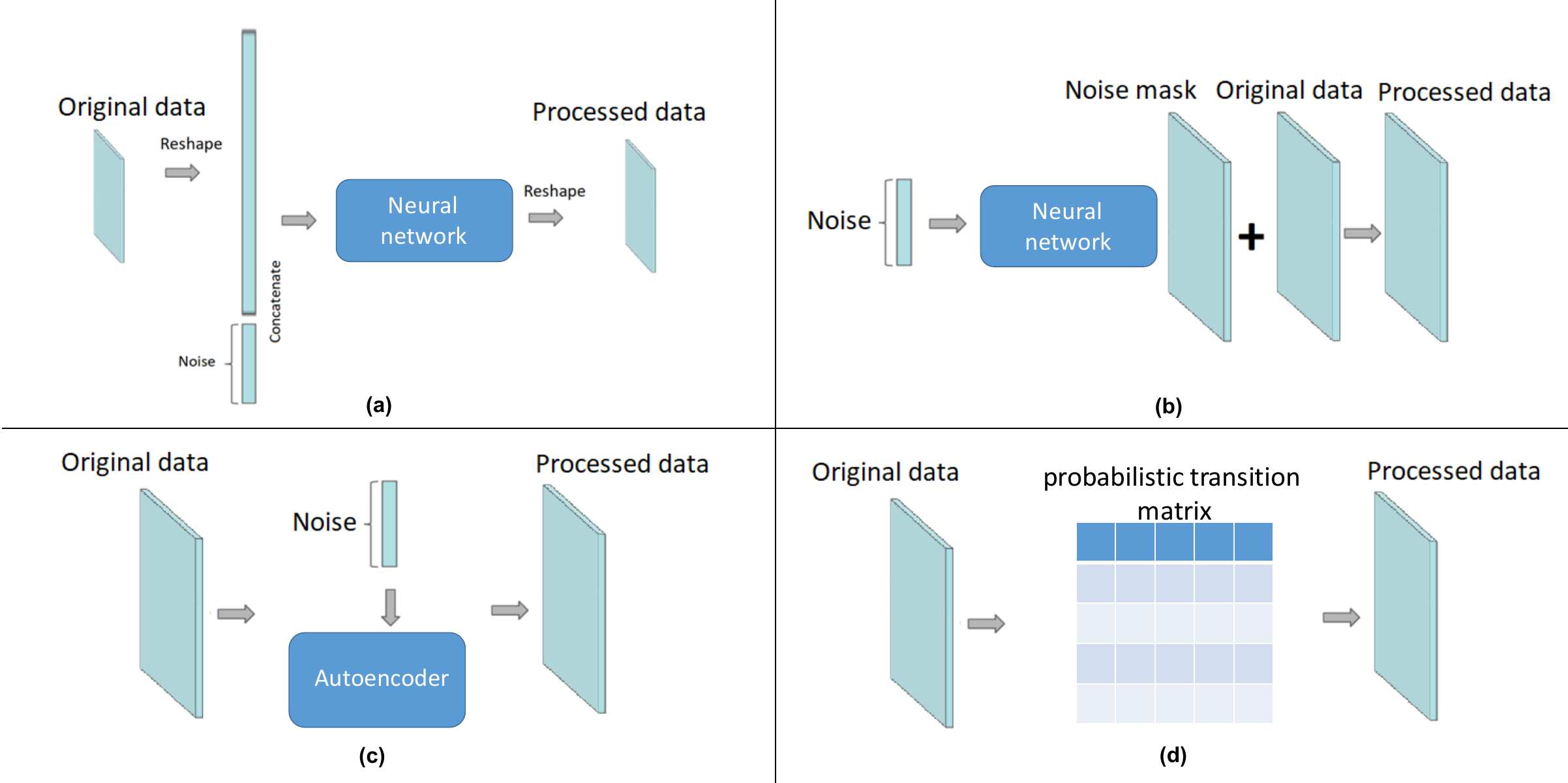}
\caption{Different architectures of the decorrelator/encoder in the FUR framework.  }
\label{fig:variousGAP}
\end{figure}

{For the four publicly available datasets, the FUR architectures we consider capture three of the four possible approaches to decorrelating the data $(X,S)$. These approaches are illustrated in Fig.~\ref{fig:variousGAP} and include: the feedforward neural network decorrelator (FNND) in Fig.~\ref{fig:variousGAP}-(a), the transposed convolution neural network decorrelator (TCNND) in Fig.~\ref{fig:variousGAP}-(b), the noisy autoencoder decorrelator (NAED) in Fig.~\ref{fig:variousGAP}-(c), and the probability matrix model (PMM) in Fig.~\ref{fig:variousGAP}-(d). 

The FNND architecture uses a feedforward multi-layer neural network to combine the low-dimensional random noise and the original data (i.e., $X$ or $(X,S)$) together. The TCNND generates high-dimensional noise from low-dimensional random noise by using a multi-layer transposed convolution neural network, and then, adds it to the original data to produce the representation $X_r$. The NAED uses the encoder of an autoencoder to generate a lower-dimensional feature vector of the original data and adds independent random noise to each element of the feature vector. The decoder of the autoencoder reconverts the noisy feature vector to generate the processed data $X_r$. Finally, for purely discrete $X$ and $S$, $X_r$ can be generated using a probability transition matrix; such a matrix can be learned from the data and is then used to map the entries of the original data to any one of the other entries using the corresponding row of the probability matrix. 

We apply both the FNND and TCNND architectures to the GENKI dataset. For the UCI Adult and HAR datasets, we use the FNND architecture, while for the UTKFace dataset, we consider the NAED architecture. Given that the datasets we consider are either continuous valued or have a mix of continuous and discrete valued features (UCI), we do not use the PMM approach in our experiments. Finally, we train our models based on the data-driven version of the FUR formulation presented in Section III of the main paper using TensorFlow \cite{abadi2016tensorflow}.

}


\section{GENKI Dataset Details and Results}\label{Subsection:GENKI}
The GENKI dataset consists of $1,740$ training and $200$ test samples. Each data sample is a $16\times16$ greyscale face image with varying facial expressions.
We choose gender 
as the sensitive $S$ and the image pixels as the non-sensitive $X$.



\subsection{Architecture}
\begin{figure}[ht]
\centering
\includegraphics[width=0.8\columnwidth]{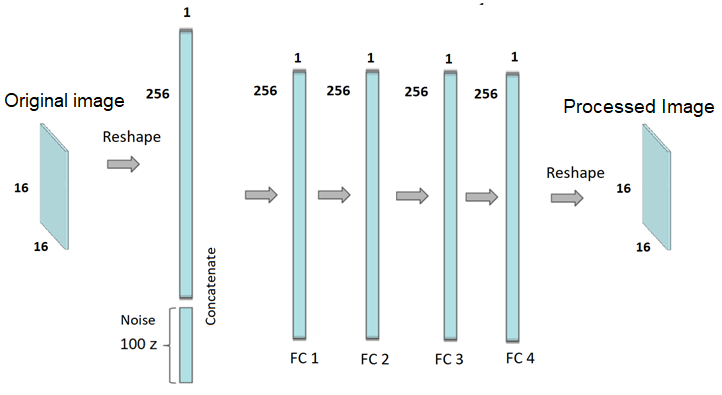}
\caption{Feedforward neural network decorrelator for the GENKI dataset.}
\label{fig:feedforwardnn}
\end{figure}

\begin{figure}[ht]
\centering
\includegraphics[width=0.8\columnwidth]{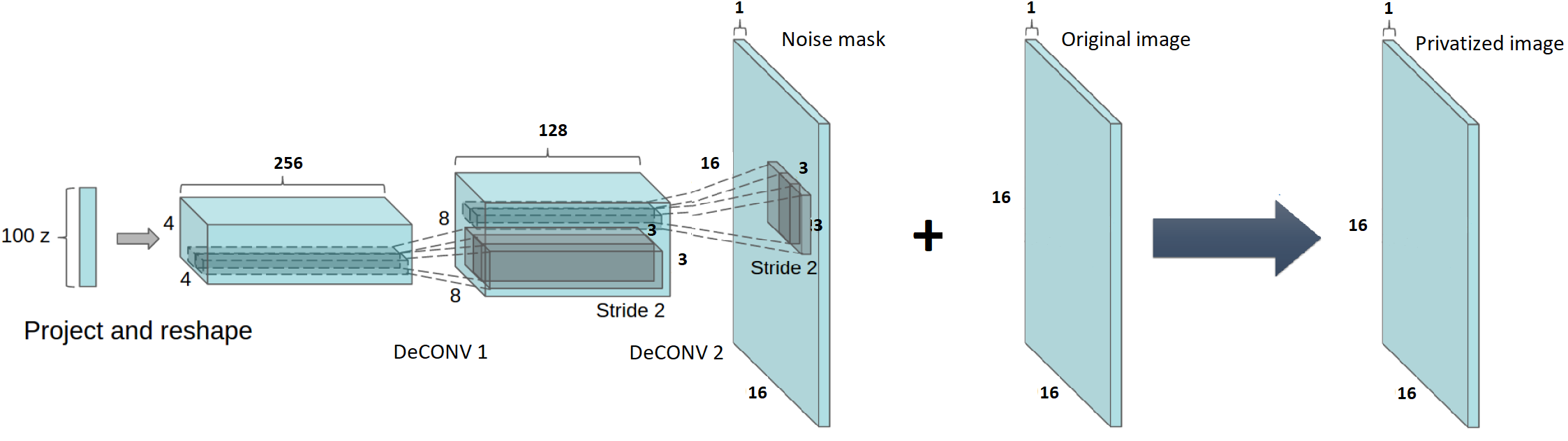}
\caption{Transposed convolution neural network decorrelator for the GENKI dataset.}
\label{fig:deconnn}
\end{figure}

\begin{figure}[ht]
\centering
\includegraphics[width=0.8\columnwidth]{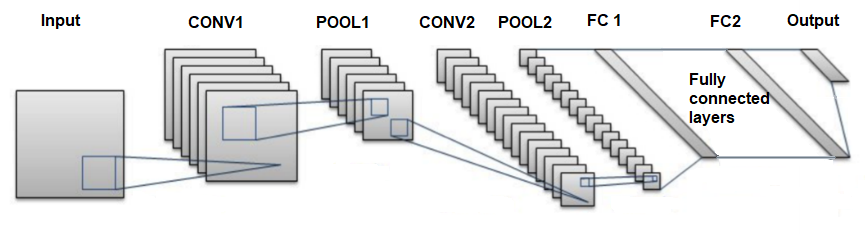}
\caption{Convolutional neural network adversary for the GENKI dataset.}
\label{fig:adversarycnn}
\end{figure}

The FNND in Fig.~\ref{fig:feedforwardnn} is modeled by a four-layer feedforward neural network. We first reshape each image into a $256\times1$ vector, and then concatenate it with a $100\times1$ Gaussian random noise vector. Each entry in the noise vector is sampled independently from a standard Gaussian distribution. We then feed the concatenated vector to a four-layer fully-connected (FC) neural network. Each layer has $256$ neurons with a leaky ReLU activation function. Finally, we reshape the output of the last layer back into a $16\times16$ image.

To model the TCNND in Fig.~\ref{fig:deconnn}, we first generate a $100\times1$ Gaussian random noise vector and use a linear projection to map it to a $4\times4\times256$ feature tensor. The feature tensor is then fed to an initial transposed convolution layer (DeCONV) with $128$ filters (filter size $3\times3$, stride 2) and a ReLU activation, followed by another DeCONV layer with $1$ filter (filter size $3\times3$, stride 2) and a tanh activation. The output of the DeCONV layer is added to the original image to generate the processed data. For both decorrelators, we add batch normalization \cite{ioffe2015batch} on each hidden layer to prevent covariance shift and help gradients to flow. We model the adversary using convolutional neural networks (CNNs), as this architecture outperforms most of the other models used for image classification \cite{krizhevsky2012imagenet,szegedy2015going}. 

Fig.~\ref{fig:adversarycnn} illustrates the architecture of the adversary. The processed images are fed to two convolution layers (CONV), CONV1 and CONV2, of size $3\times3\times32$ and $3\times3\times64$, respectively. Each convolution layer is followed by rectified linear unit (ReLU) activation and batch normalization. The output of each convolution layer is fed to a $2\times2$ maxpool layer (POOL) to extract features for classification. The second maxpool layer is followed by two fully-connected layers, each containing $1024$ neurons with batch normalization and ReLU activation. Finally, the output of the last fully-connected layer is mapped to the output layer, which contains two neurons that capture the belief of the subject being a male or a female.

We note that the architecture for the GENKI expression classifier mirrors that of the adversary since it also involves taking the learned representation $X_r$ and reducing it to an appropriate feature space to finally perform binary classification. The key difference is that now the two neurons in the last layer capture the belief of the subject smiling or not smiling.

\subsection{Computing Differential Censoring Risk}
In the local differential privacy setting, we consider the value of each pixel as the output of the query function. Since the pixel values of the GENKI dataset are normalized between 0 and 1, we can use the dimension of the image to bound the sensitivities for both the Laplace and Gaussian mechanisms. For the Laplace mechanism, let $X$ and $X'$ be two different image vectors. Since the value of $X$ and $X'$ in each dimension is between $0$ and $1$, $|X_i - X'_i| \le 1$ for the $i^{\text{th}}$ dimension. Thus, the $L_1$ sensitivity can be bounded by the dimension of the image $d$. Notice that the variance of the Laplace noise parameterized by $\lambda$ is $2\lambda^2$. Since we are adding independent Laplace noise to each pixel, given the per image distortion $D$, we have $$\frac{D}{d} = 2\frac{d^2}{\epsilon^2} \implies \epsilon = d\sqrt{\frac{2d}{D}}.$$
Similarly, for the Gaussian mechanism, we can bound the $L_2$ sensitivity by $\sqrt{d}$. For a fixed $\delta$, the privacy risk $\epsilon$ for adding Gaussian noise with variance $\sigma^2$ is given by $$\epsilon = 2\frac{\sqrt{d}}{\sigma} \sqrt{\ln\left(\frac{1.25}{\delta}\right)}.$$ Since we are adding Gaussian noise independently to each pixel, the variance of the per pixel noise in given by $\sigma = \frac{D}{d}$. As a result, we have $$\epsilon = 2\frac{{d}}{\sqrt{D}} \sqrt{\ln\left(\frac{1.25}{\delta}\right)}. $$ 

\subsection{Architecture for Mutual Information Estimation}
\label{MIestimation}
For the GENKI dataset, as the first step towards estimating MI, we need to extract the feature vector $X_f$. To do so, we employ an architecture which is similar to those used for the adversary (and classifier); however, in the last layer, we reduce the features to a significantly lower dimensional space. We present these details below.

The network which extracts the low dimensional $\hat{X}_f$ vector involves the following: a CNN composed of two convolutional (conv) blocks (described below), followed by two fully-connected (FC) layers, and ending with a two neuron-layer with softmax activation. In each conv block, we have a convolution layer that uses $3 \times 3$ filters 
and a stride of $1$, a $2\times 2$ max-pooling layer with a stride of $2$, and a ReLU activation function. The first and second conv blocks have 32 and 64 filters, respectively. We flatten out the output of the second conv block, yielding a 256-dimensional vector. This vector, which results from extracting features from the second conv block, is passed through the first FC layer with batch normalization and a ReLU activation to get an 8-dimensional vector. This 8-dimensional vector serves as an input to a second FC layer that outputs a 2-dimensional vector to which the softmax function is applied. The aforementioned 8-dimensional vector is the feature embedding vector $\hat{X}_f$ in our empirical MI estimation.

The above network is trained first using $X_r$ and $S$ to estimate the sensitive feature $S$ as $\hat{S}$ (as the adversary would); the resulting 8-dimensional vector prior to the final layer described above is the $\hat{X}_g$ used to compute $\hat{I}(\hat{X}_g;S)$. We then train the model in a similar fashion using $X_r$ and $Y$ data to obtain $\hat{X}_f$ and thereby compute $\hat{I}(\hat{X}_f;Y)$ as an estimate of the MI between $X_r$ and $Y$.

\section{HAR Dataset Details and Results}\label{Subsection:HAR}
 
The HAR dataset consists of $561$ features of motion sensor data collected by a smartphone from 30 subjects performing six activities (walking, walking upstairs, walking downstairs, sitting, standing, laying). Each feature is normalized between $-1$ and $1$. We choose subject identity as the sensitive $S$ and the features of motion sensor data as the non-sensitive $X$.


\subsection{Architecture}
We first concatenate the original data with a $100\times1$ Gaussian random noise vector. We then feed the entire $661 \times 1$ vector to a feedforward neural network with three hidden fully-connected (FC) layers. Each hidden layer has $512$ neurons with leaky ReLU activation. Finally, we use another FC layer with $561$ neurons to generate the processed data. 

For the adversary, we use a five-layer feedforward neural network. The hidden layers 1 to 5 have $512$, $512$, $256$, and $128$ neurons with leaky ReLU activation, respectively. The output of the last hidden layer is mapped to the output layer, which contains $30$ neurons that capture the belief of the subject's identity. For both the decorrelator and the adversary, we add batch normalization after the output of each hidden layer.

The architecture for the classifier largely mimics that of the adversary except now we wish to predict one of the six activity labels from the representation fed to the classifier. Thus, we again use a five-layer feedforward neural network with the hidden layers having $512$, $512$, $256$, and $128$ neurons with leaky ReLU activation, respectively. The output of the last hidden layer is mapped to the softmax output layer, which now contains $6$ neurons that capture the belief for each of the six activities. Here again, we add batch normalization after the output of each hidden layer.

\subsection{Architecture for Mutual Information Estimation}
\begin{figure}[ht]
\centering
\includegraphics[width=0.8\columnwidth]{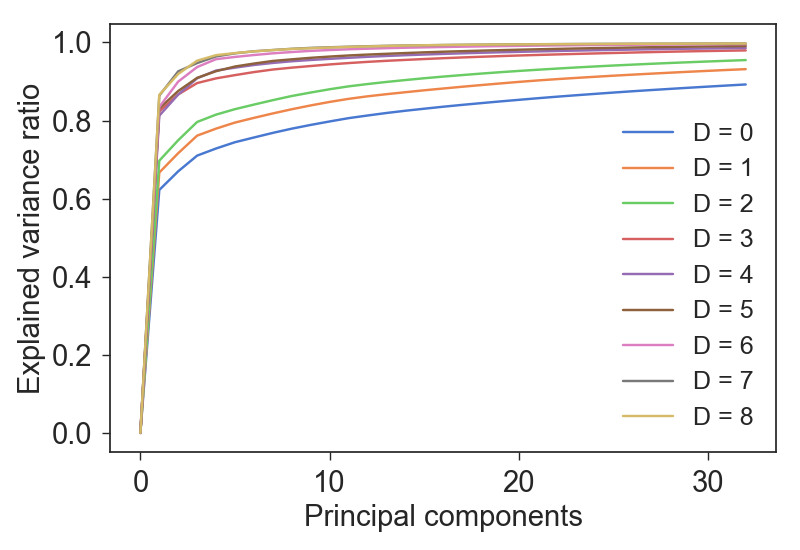}
\caption{Top 32 principal components out of the 561 features with different distortion $D$ for processed data in HAR.}
\label{fig:har:pca:top32}
\end{figure}

The same general model for mutual information estimation as that described in Section \ref{MIestimation} is applied here. However, compared to the GENKI dataset, estimating mutual information for the HAR dataset presents a slightly different challenge, as the size of the alphabet for the sensitive label (i.e. identity) is 30. Thus, it requires at least 30 neurons prior to the output layer of the corresponding classification task. In fact, we pose 128 neurons before the final softmax output layer in order to get a reasonably good classification accuracy. Using the 128-dimensional vector as our feature embedding to calculate mutual information is almost impossible due to the curse of dimensionality. Therefore, we apply Principal Component Analysis (PCA), shown in Fig.~\ref{fig:har:pca:top32}, and pick the first 12 components to circumvent this issue. The resulting 12-dimensional vector is considered to be an approximate feature embedding that encapsulates the major information of the processed data.    

\section{UCI Adult Dataset Details}\label{Subsection:UCI}
Each sample in the UCI Adult dataset has both continuous and categorical features. Table \ref{tb:UCIAdult_dataset} lists all the considered features. We perform a one-hot encoding on each categorical feature in $(S,X)$ and store the mapping function from the one-hot encoding to the categorical data. We restrict the continuous features in $X$ to the interval $(0,1)$ using normalization. 

\begin{table*}[]
\centering
\small
\begin{tabular}{|c|l|l|c|}
\hline
Case I  & Feature        &  Description  & Case II \\ \hline
$Y$  & salary        & $2$-salary intervals: $>50K$ and $\leq 50K$  & $Y$ \\\hline
$S$ &  gender      & $2$ classes: male and female & \multirow{2}{*}{\centering $S$} \\ \cline{1-3}
\multirow{12}{*}{\centering $X$}  & relationship      & $6$  classes of family relationships (e.g., wife and  husband) & \\ \cline{2-4}
  & age           & $9$-age intervals: $18-25$, $25-30$, $30-35$, ...,$60-65$  & \multirow{11}{*}{\centering $X$} \\ \cline{2-3}
& workclass     & $8$ types of employer + unknown  & \\ \cline{2-3}
& education     & $16$ levels of the highest achieved education & \\ \cline{2-3}
& marital-status  & $7$ classes of marital status  & \\ \cline{2-3}
& occupation       & $14$ types of occupation + unknown   & \\ \cline{2-3}
& race          & $5$ classes  & \\ \cline{2-3}
& native-country     & $41$ countries of origin + unknown & \\ \cline{2-3}
& capital-gain      & Recorded capital gain; (continuous)     & \\ \cline{2-3}
& capital-loss      &  Recorded capital loss; (continuous)      & \\ \cline{2-3}
& hours-per-week    &  Worked hours per week; (continuous) & \\ \cline{2-3}
& education-num     & Numerical version of education; (continuous)     &       \\ \hline
\end{tabular}
\caption{Features of the UCI Adult dataset.}
\label{tb:UCIAdult_dataset}
\end{table*}

\begin{figure}[!hbpt]
	\centering
	\begin{subfigure}{0.48\textwidth}
      \includegraphics[width=0.9\columnwidth]{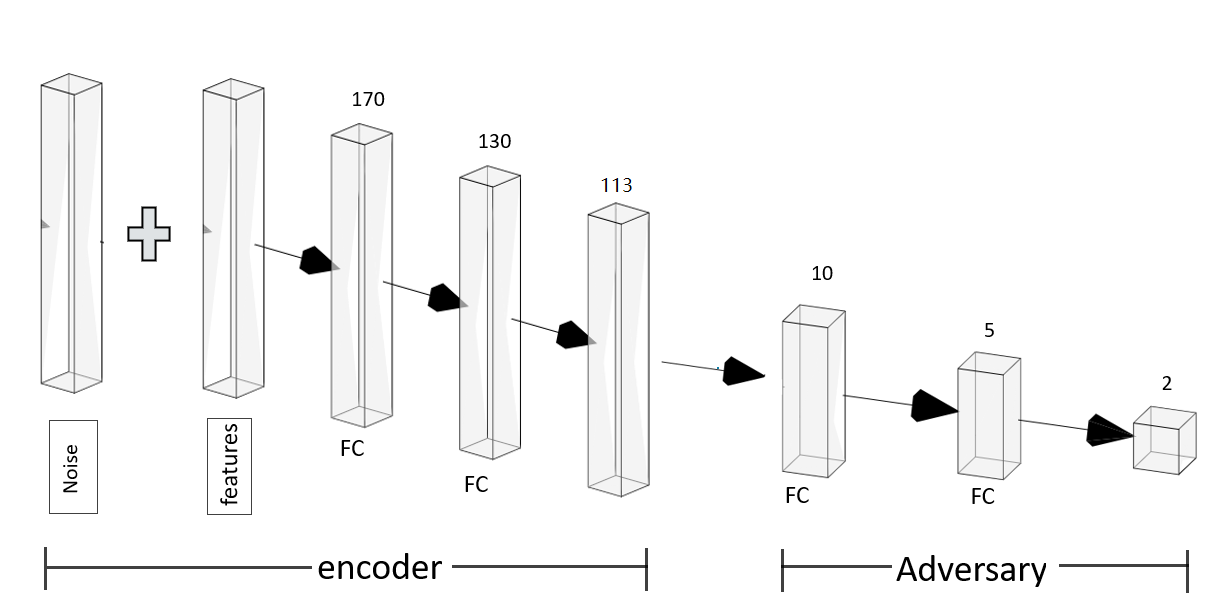}
		\caption{UCI Dataset: The architecture for Case I ($S$ is the binary gender variable) }
		\label{fig:UCI_Architecture-CaseI}
	\end{subfigure}
	\begin{subfigure}{0.48\textwidth}
\includegraphics[width=0.9\columnwidth]{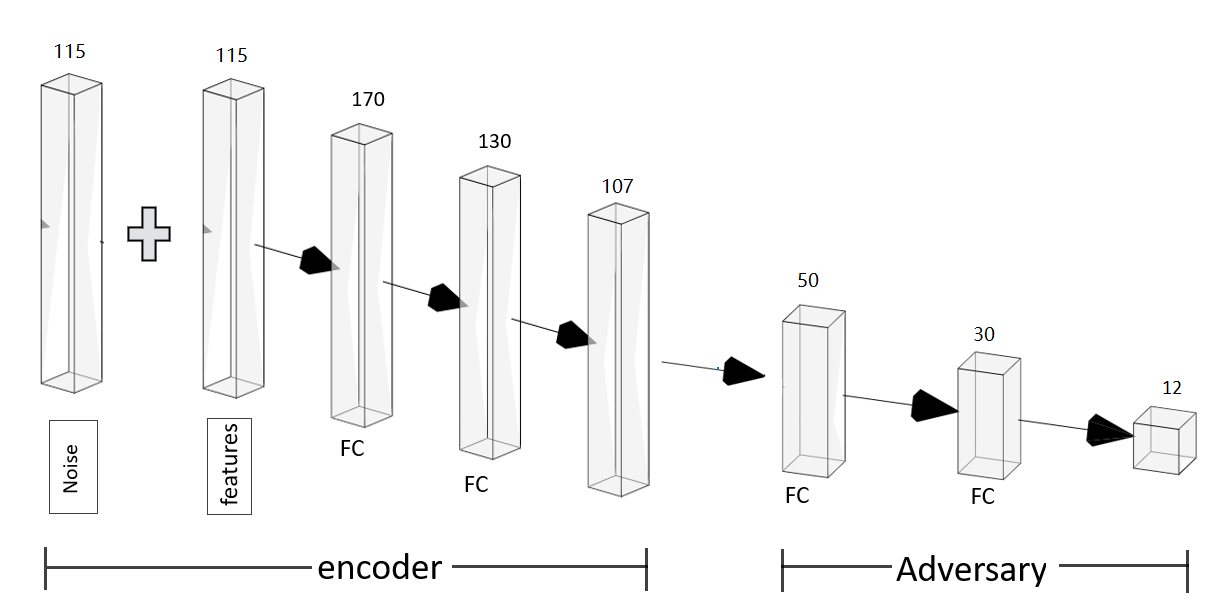}
		\caption{UCI Dataset: The architecture for Case II ($S$ is non-binary and includes both gender and relationship) }
\label{fig:UCI_Architecture-CaseII}
	\end{subfigure}
		\caption{The architectures of the encoder and adversary for the UCI Adult dataset for both Cases I and II. } 
	\label{fig:UCI_Architecture}
\end{figure}

\subsection{Architecture}
The two architectures used are shown in Fig.~\ref{fig:UCI_Architecture}.
We concatenate the pre-processed data with a same-size standard Gaussian random vector and feed the entire vector to the encoder. The encoder consists of two fully-connected (FC) hidden layers, the first with $170$ neurons and the second with $130$ neurons. Since the output representation $X_r$ has the same dimension as the feature variable $X$, the output layer of the encoder has $113$ (as shown in Fig.~\ref{fig:UCI_Architecture-CaseI}) and $107$ (as shown in Fig.~\ref{fig:UCI_Architecture-CaseII}) neurons for Case I and Case II, respectively. We use a ReLU activation function in the encoder. 

We recall the two cases considered for this dataset: Case I with binary sensitive feature $S$ (gender) and Case II with non-binary $S$ (gender and relationship). For Case I, the inputs can be either $X$ only or both $X$ and $S$. With only $X$ as input to the encoder, the length of the input vector is $226$, and when both $X$ and $S$ (binary) are inputs, the input vector length is $230$. In both scenarios, the length of the encoder's output vector is $113$. For Case II, since both $X$ and $S$ are inputs, the length of the input vector is $230$. The length of the encoder's output vector is $107$, since $S$ is non-binary in this case.

For Case I, the adversarial classifier in Fig.~\ref{fig:UCI_Architecture-CaseI} consists of three fully-connected (FC) layers 1 to 3 with $10$, $5$ and $2$ neurons, respectively, and it takes $X_r$ as the input and outputs a probability distribution for the binary sensitive variable $S$ (i.e., gender). Here, ReLU is used as the activation function in the two hidden layers and soft-max is used in the output layer to generate the probability distribution for gender. The same architecture is used for the downstream application of salary classification. For Case II, the adversarial classifier in Fig.~\ref{fig:UCI_Architecture-CaseII} consists of three fully-connected (FC) layers 1 to 3 with $50$, $30$ and $12$ neurons, respectively. Leaky ReLU is used as the activation function in the two hidden layers and soft-max is used in the output layer.
All of the above models use log-loss as the loss function and are optimized using Adam optimizer \cite{kingma2017adam}.

\section{UTKFace Dataset Details}\label{Subsection:UTK}
The UTKFace dataset \cite{zhifei2017cvpr} 
consists of more than $20$k $200\times 200$ color images of faces labeled by age, ethnicity, and gender. Individuals in the dataset have ages from $0$ to $116$ years and are divided into $5$ ethnicities: White, Black, Asian, Indian, and others including Hispanic, Latino and Middle Eastern.
We take gender as $S$, the image as $X$, and age and ethnicity as two target labels $Y$.
We reshape the original $200\times 200$ color images (of faces) in the UTKFace dataset to color images of size  $64\times 64$ and we restrict the data to contain images for ages between 10 and 65.
\begin{figure}[!hbpt]
	\centering
      \includegraphics[width=1\columnwidth]{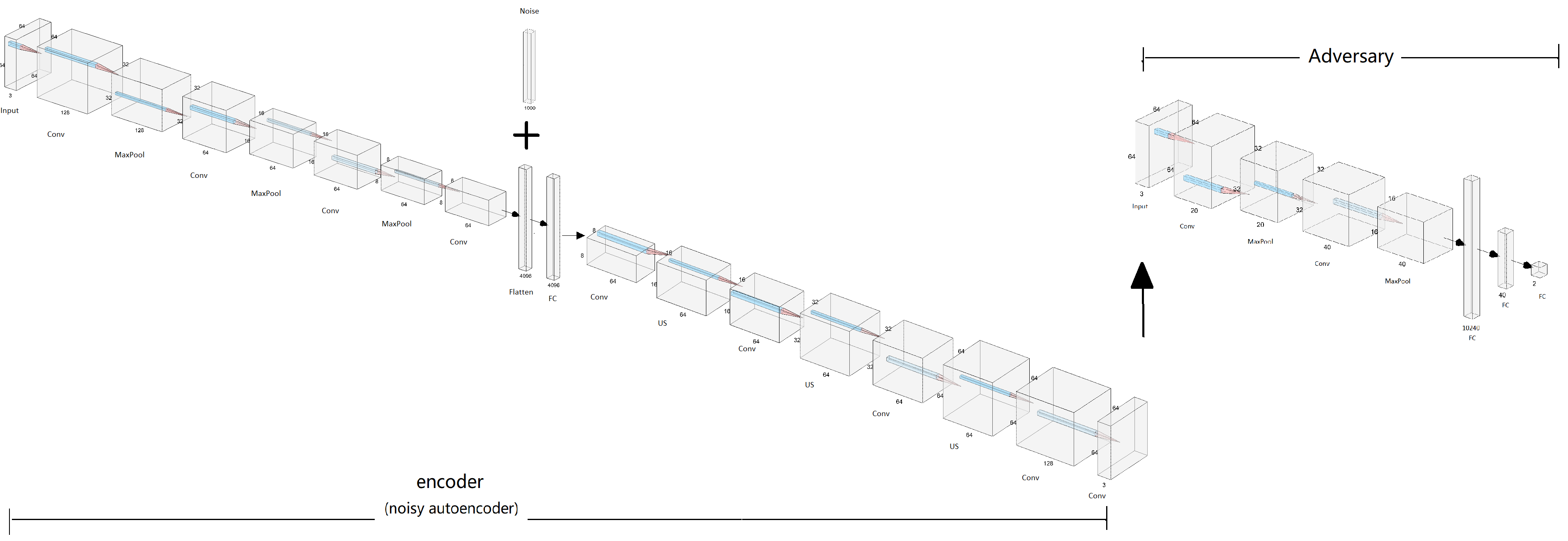}
	\caption{The architectures of the encoder and adversary for the UTKFace dataset.}%
	\label{fig:Architectures_UTKFace}
\end{figure}

\begin{figure}[!hbpt]
	\centering
      \includegraphics[width=0.9\columnwidth]{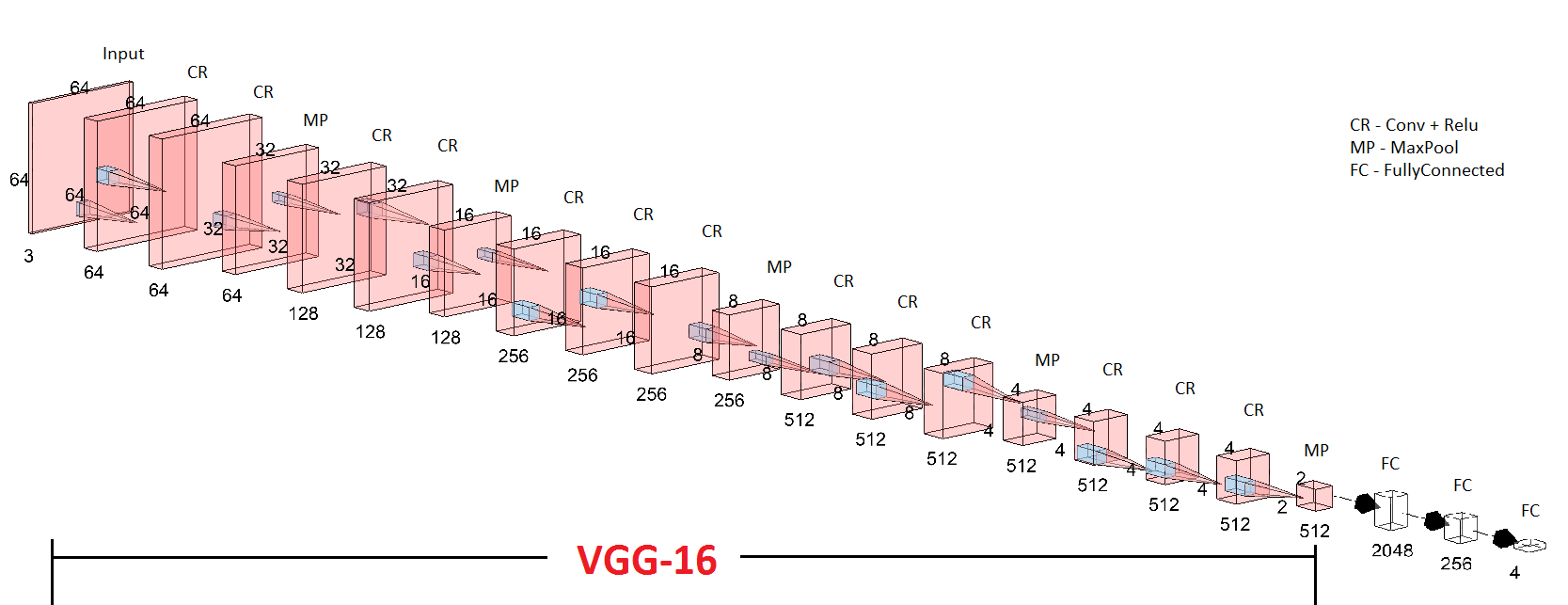}
        \caption{The architecture of the neural network for ethnicity classification for the UTKFace dataset.}
      \label{fig:Architectures_UTKFace_race}
\end{figure}

\begin{figure}[!hbpt]
    \centering
    \includegraphics[width=0.9\columnwidth]{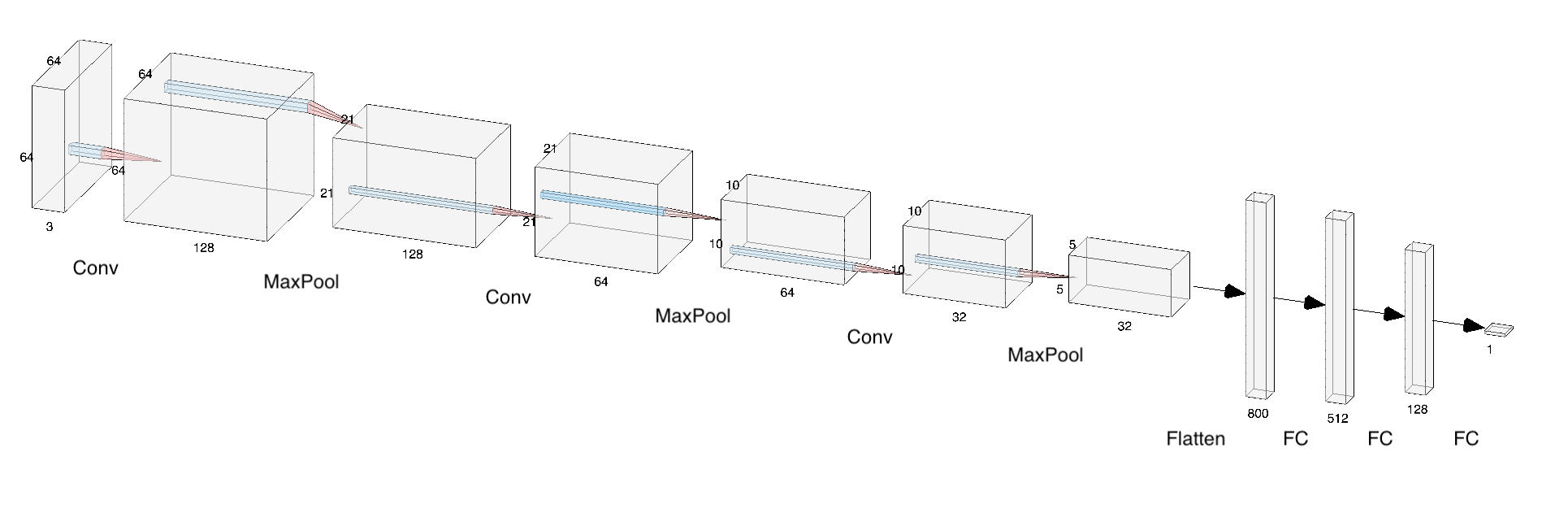}
        \caption{The architecture of the neural network for age regression for the UTKFace dataset.}
    \label{fig:Architectures_UTKFace_age}
\end{figure}

\subsection{Architecture}
The architectures used are shown in Figs.~\ref{fig:Architectures_UTKFace}, \ref{fig:Architectures_UTKFace_race}, and \ref{fig:Architectures_UTKFace_age}. Fig.~\ref{fig:Architectures_UTKFace} illustrates the architecture of the FUR model, which consists of an encoder and an adversarial classifier. The encoder is implemented using a noisy autoencoder, whose encoder transforms the original $64\times64$ RGB-images into a $4096$-dimensional feature vector. This feature vector is mixed with a $4096$-dimensional standard normal random vector\footnote{A random vector is a standard normal random vector if all of its components are independent and identically distributed following the standard normal distribution.} and then fed into a decoder to reconstruct a $64\times 64$ colorful image, which is the universal representation $X_r$. This differs from a standard autoencoder, in which the feature vector is directly fed into the decoder. Specifically, the encoder part of the noisy autoencoder consists of four convolution layers 1 to 4 with $128$, $64$, $64$ and $64$ output channels, respectively, and three $2\times 2$-max pooling layers following the first three convolution layers. The encoder part is followed by two fully-connected layers, each with $4096$ neurons, which mix the noise and the output feature vector. 
The following decoder part  
consists of five convolution layers 1-5 with $64$, $64$, $64$, $128$ and $3$ output channels, respectively, and three $2\times 2$-up-sampling layers following the first three convolution layers. The adversarial classifier takes in the representation $X_r$ and outputs the prediction of the sensitive $S$ (gender). It consists of two convolution layers, the first with $20$ output channels and the second with $40$ output channels, two $2\times 2$-max pooling layers following each of the convolution layers, and two fully-connected layers, the first with $40$ neurons and the second with $2$ neurons.
The kernels in the convolution layers are of size $3\times 3$. All convolution and fully-connected layers use ReLU as the activation function, except the last layers of the decoder and the adversary, which use sigmoid and softmax, respectively. The encoder and adversarial classifier use the square-loss and log-loss as the loss functions, respectively, and both are optimized using the Adam optimizer.

Fig.~\ref{fig:Architectures_UTKFace_race} illustrates the architecture of the downstream non-binary classifier for ethnicity. The classifier is built by changing the top (last) $3$ fully-connected layers of the VGG $16$ model\footnote{https://keras.io/applications/\#vgg16} pre-trained on ImageNet. The first layer has $256$ neurons with ReLU as the activation function and is followed by a dropout layer with the rate $0.5$; the second layer has $4$ neurons with softmax as the activation function. The classifier uses log-loss and is optimized by a stochastic gradient descent optimizer.

Fig.~\ref{fig:Architectures_UTKFace_age} illustrates the architecture of the neural network used in the downstream application of age regression. The neural network consists of three $3\times 3$ convolution layers 1 to 3 with $128$, $64$ and $32$ output channels, respectively, three $2\times 2$-max pooling layers following each of the convolution layers, and three fully-connected layers 1 to 3 with $512$, $128$ and $1$ neurons, respectively. All layers use ReLU as the activation function, except the last layer, which uses linear activation. The model uses the squared loss as the loss function and is optimized using Adam optimizer. 

\bibliographystyle{IEEEtran}
\bibliography{Bib-Files}